\documentclass[11pt,a4paper]{article}
\pdftrailerid{}
\PassOptionsToPackage{table}{xcolor}
\usepackage[preprint]{acl}
\usepackage[T1]{fontenc}
\usepackage{times}
\usepackage{graphicx}
\usepackage{booktabs}
\usepackage{array}
\usepackage{tabularx}
\usepackage{multirow}
\usepackage{xcolor}
\usepackage{pifont}
\usepackage{amsmath}
\usepackage{amssymb}
\usepackage{tikz}

\newif\ifpapersynthetic
\papersyntheticfalse
\newif\ifpapercoded
\papercodedfalse
\newif\ifpaperheld
\paperheldfalse
\newcommand{\num}[2][]{%
  \ifcsname paper@num@#1@#2\endcsname\csname paper@num@#1@#2\endcsname
  \else\PackageError{paper}{Undefined number [#1]{\detokenize{#2}}}{Regenerate the paper assets.}\fi}
\expandafter\def\csname paper@num@a2@facts.own_metric.knn_router.points_without_two_largest_gains\endcsname{0.57}
\expandafter\def\csname paper@num@a3@fu.pool.top_k_minus_stacking_log_loss\endcsname{0.013}
\expandafter\def\csname paper@num@a3@fu.unknown.across_capped_minus_text_log_loss\endcsname{0.019}
\expandafter\def\csname paper@num@a3@ops.suites.main35.contrasts.router_uncapped_vs_control.log_loss.difference\endcsname{0.010}
\expandafter\def\csname paper@num@a4@fu.full_text.ridge_minus_best_log_loss.difference\endcsname{0.0057}
\expandafter\def\csname paper@num@a4@fu.full_text.ridge_minus_best_log_loss_no_recall_only.difference\endcsname{0.0025}
\expandafter\def\csname paper@num@f0@facts.models.cost_ratio\endcsname{22}
\expandafter\def\csname paper@num@f0@facts.over_benchmarks_summary.widest_ratio\endcsname{43}
\expandafter\def\csname paper@num@f0@facts.relative_to_best_single.uniform_mean.cost_ratio\endcsname{12}
\expandafter\def\csname paper@num@f1@diag.suites.main35.overall.headroom.0.effective_voters\endcsname{2.2}
\expandafter\def\csname paper@num@f1@facts.homogeneity.effective_voters_max\endcsname{3.3}
\expandafter\def\csname paper@num@f1@facts.homogeneity.effective_voters_min\endcsname{1.4}
\expandafter\def\csname paper@num@f1@facts.over_benchmarks_summary.narrowest_ratio\endcsname{1.5}
\expandafter\def\csname paper@num@f1@facts.relative_to_best_single.greedy.cost_ratio\endcsname{1.9}
\expandafter\def\csname paper@num@f1@facts.relative_to_best_single.stack_final_temperature.cost_ratio\endcsname{3.4}
\expandafter\def\csname paper@num@f1@facts.restyle.latency.largest_ratio\endcsname{1.7}
\expandafter\def\csname paper@num@f1@facts.restyle.routing.reference_cost_ratio_to_top\endcsname{2.1}
\expandafter\def\csname paper@num@f1@facts.restyle.routing.reference_points_over_top\endcsname{4.4}
\expandafter\def\csname paper@num@f1@facts.restyle.routing.router_points_over_top\endcsname{3.6}
\expandafter\def\csname paper@num@f1@fu.pool.top_k_calls\endcsname{3.0}
\expandafter\def\csname paper@num@f1@homog.suites.main35.all_pairs.joint_failure_ratio\endcsname{5.7}
\expandafter\def\csname paper@num@f1@ops.suites.main35.points.confidence_cascade.uncapped.calls_per_request\endcsname{2.6}
\expandafter\def\csname paper@num@f2@diag.suites.main35.overall.across_models.pairs.0.error_correlation\endcsname{0.45}
\expandafter\def\csname paper@num@f2@facts.models.spearman_accuracy_cost\endcsname{0.01}
\expandafter\def\csname paper@num@f2@facts.recall_only.stack_final_temperature.headline\endcsname{0.96}
\expandafter\def\csname paper@num@f2@facts.recall_only.stack_final_temperature.without\endcsname{1.03}
\expandafter\def\csname paper@num@f2@facts.restyle.all_pairs_error_correlation\endcsname{0.45}
\expandafter\def\csname paper@num@f2@facts.restyle.latency_accuracy_spearman\endcsname{0.13}
\expandafter\def\csname paper@num@f2@facts.restyle.top_models_error_correlation\endcsname{0.57}
\expandafter\def\csname paper@num@f2@fu.few_right.confidence_right\endcsname{0.55}
\expandafter\def\csname paper@num@f2@fu.few_right.confidence_wrong\endcsname{0.59}
\expandafter\def\csname paper@num@f2@homog.suites.main35.all_pairs.error_correlation\endcsname{0.45}
\expandafter\def\csname paper@num@f2@homog.suites.main35.all_pairs.error_correlation_given_others\endcsname{0.06}
\expandafter\def\csname paper@num@f2@homog.suites.main35.different_organisation.error_correlation_given_others\endcsname{0.04}
\expandafter\def\csname paper@num@f2@homog.suites.main35.same_organisation.error_correlation_given_others\endcsname{0.17}
\expandafter\def\csname paper@num@f2@ops.suites.main35.points.confidence_cascade.uncapped.calls_per_request\endcsname{2.62}
\expandafter\def\csname paper@num@f2@rows.headline.certified_exit.latency_mean_s\endcsname{1.32}
\expandafter\def\csname paper@num@f2@rows.headline.stack_final_temperature.latency_mean_s\endcsname{1.21}
\expandafter\def\csname paper@num@f3@facts.calibration_order.calibrated_members\endcsname{0.481}
\expandafter\def\csname paper@num@f3@facts.calibration_order.calibrated_members_and_final_temperature\endcsname{0.417}
\expandafter\def\csname paper@num@f3@facts.calibration_order.final_temperature\endcsname{0.426}
\expandafter\def\csname paper@num@f3@facts.calibration_order.none\endcsname{0.494}
\expandafter\def\csname paper@num@f3@facts.floor.largest_shift_of_other_rows\endcsname{0.002}
\expandafter\def\csname paper@num@f3@facts.no_target_labels.transfer_single\endcsname{0.550}
\expandafter\def\csname paper@num@f3@facts.restyle.representations.log_loss_spread\endcsname{0.006}
\expandafter\def\csname paper@num@f3@folds.suites.main35.largest_log_loss_shift\endcsname{0.015}
\expandafter\def\csname paper@num@f3@rows.headline.majority_vote.log_loss\endcsname{1.708}
\expandafter\def\csname paper@num@f3@rows.headline.majority_vote.log_loss_floor_1e-3\endcsname{0.592}
\expandafter\def\csname paper@num@f3@rows.headline.random_single.log_loss\endcsname{0.671}
\expandafter\def\csname paper@num@f3@rows.headline.random_single.log_loss_floor_1e-3\endcsname{0.627}
\expandafter\def\csname paper@num@n@audit.corpus.adapted_observations\endcsname{5,286}
\expandafter\def\csname paper@num@n@audit.corpus.attempts\endcsname{1,837,835}
\expandafter\def\csname paper@num@n@audit.corpus.benchmarks\endcsname{35}
\expandafter\def\csname paper@num@n@audit.corpus.fields\endcsname{267,716}
\expandafter\def\csname paper@num@n@audit.corpus.judged_fields\endcsname{225,391}
\expandafter\def\csname paper@num@n@audit.corpus.models\endcsname{17}
\expandafter\def\csname paper@num@n@audit.corpus.models_answering_in_smaller_batches\endcsname{2}
\expandafter\def\csname paper@num@n@audit.corpus.options_max\endcsname{151}
\expandafter\def\csname paper@num@n@audit.corpus.options_min\endcsname{2}
\expandafter\def\csname paper@num@n@audit.corpus.organisations\endcsname{6}
\expandafter\def\csname paper@num@n@audit.corpus.requests\endcsname{106,845}
\expandafter\def\csname paper@num@n@audit.corpus.structure.per_item_inventory\endcsname{18}
\expandafter\def\csname paper@num@n@audit.corpus.successful_physical_calls\endcsname{1,827,113}
\expandafter\def\csname paper@num@n@audit.corpus.vectors\endcsname{4,551,172}
\expandafter\def\csname paper@num@n@audit.rights.benchmark_source_category_counts.clear\endcsname{15}
\expandafter\def\csname paper@num@n@audit.rights.benchmark_source_category_counts.restricted\endcsname{12}
\expandafter\def\csname paper@num@n@audit.rights.benchmark_source_category_counts.unclear\endcsname{8}
\expandafter\def\csname paper@num@n@audit.rights.model_status_counts.cleared\endcsname{2}
\expandafter\def\csname paper@num@n@audit.rights.model_status_counts.conditional\endcsname{11}
\expandafter\def\csname paper@num@n@audit.rights.model_status_counts.unverified\endcsname{4}
\expandafter\def\csname paper@num@n@audit.telemetry.price_card_models_exact\endcsname{16}
\expandafter\def\csname paper@num@n@audit.telemetry.price_card_rows_beyond_rounding\endcsname{34}
\expandafter\def\csname paper@num@n@audit.vectors.models_with_zero_entries\endcsname{5}
\expandafter\def\csname paper@num@n@audit.vectors.reported_differs_first\endcsname{971}
\expandafter\def\csname paper@num@n@audit.vectors.zero_mass_on_accepted\endcsname{3,689}
\expandafter\def\csname paper@num@n@det.files_compared\endcsname{35}
\expandafter\def\csname paper@num@n@det.reruns\endcsname{2}
\expandafter\def\csname paper@num@n@facts.bootstrap.draws\endcsname{2,000}
\expandafter\def\csname paper@num@n@facts.floor.rows_moving_more_than_0_003\endcsname{2}
\expandafter\def\csname paper@num@n@facts.label_budget.groups_where_selection_beats_transfer\endcsname{12}
\expandafter\def\csname paper@num@n@facts.panel_size.greedy.first_size_within_0_001\endcsname{8}
\expandafter\def\csname paper@num@n@facts.panel_size.ranked_top_k.best_size\endcsname{2}
\expandafter\def\csname paper@num@n@facts.planned_contrasts.computed\endcsname{7}
\expandafter\def\csname paper@num@n@facts.restyle.complementarity.benchmark_winners\endcsname{10}
\expandafter\def\csname paper@num@n@facts.restyle.complementarity.models\endcsname{17}
\expandafter\def\csname paper@num@n@facts.restyle.representations.count\endcsname{4}
\expandafter\def\csname paper@num@n@facts.sensitivity.other_suites\endcsname{3}
\expandafter\def\csname paper@num@n@folds.draws\endcsname{20}
\expandafter\def\csname paper@num@n@folds.suites.main35.contrasts_compared\endcsname{10}
\expandafter\def\csname paper@num@n@fu.control.cells\endcsname{175}
\expandafter\def\csname paper@num@n@fu.control.cells_one_model\endcsname{70}
\expandafter\def\csname paper@num@n@fu.control.log_loss_better\endcsname{10}
\expandafter\def\csname paper@num@n@fu.control.log_loss_worse\endcsname{22}
\expandafter\def\csname paper@num@n@fu.full_text.stored_below_half\endcsname{7}
\expandafter\def\csname paper@num@n@fu.full_text.stored_one_vector\endcsname{3}
\expandafter\def\csname paper@num@n@fu.own.router_above\endcsname{10}
\expandafter\def\csname paper@num@n@fu.own.router_below\endcsname{25}
\expandafter\def\csname paper@num@n@fu.own.stack_above\endcsname{31}
\expandafter\def\csname paper@num@n@fu.own.stack_above_own\endcsname{30}
\expandafter\def\csname paper@num@n@fu.own.stack_below\endcsname{2}
\expandafter\def\csname paper@num@n@fu.own.stack_below_own\endcsname{4}
\expandafter\def\csname paper@num@n@fu.ridge.variants\endcsname{4}
\expandafter\def\csname paper@num@n@fu.ridge.whole_request_variants\endcsname{2}
\expandafter\def\csname paper@num@n@homog.suites.main35.all_models.none_right_with_choice\endcsname{4,220}
\expandafter\def\csname paper@num@n@homog.suites.main35.all_pairs.both_wrong_agreement_benchmarks\endcsname{21}
\expandafter\def\csname paper@num@n@leak.benchmarks_affected\endcsname{5}
\expandafter\def\csname paper@num@n@leak.requests_with_duplicate_text_across_folds\endcsname{23}
\expandafter\def\csname paper@num@n@ops.suites.main35.features.at_most_one\endcsname{21}
\expandafter\def\csname paper@num@n@ops.suites.main35.features.benchmarks\endcsname{35}
\expandafter\def\csname paper@num@n@ops.suites.main35.selection.confidence_cascade.cap_changed_choice\endcsname{144}
\expandafter\def\csname paper@num@n@ops.suites.main35.selection.confidence_cascade.cells\endcsname{175}
\expandafter\def\csname paper@num@n@ops.suites.main35.selection.knn_router.cap_changed_choice\endcsname{71}
\expandafter\def\csname paper@num@n@ops.suites.main35.selection.knn_router.cells\endcsname{175}
\expandafter\def\csname paper@num@n@ops.suites.main35.selection.knn_router.none_within_budget\endcsname{17}
\expandafter\def\csname paper@num@n@rel.excluded_request_benchmarks\endcsname{2}
\expandafter\def\csname paper@num@n@rel.excluded_requests\endcsname{41}
\expandafter\def\csname paper@num@n@rel.models_held\endcsname{5}
\expandafter\def\csname paper@num@n@rel.secondary_host_cells_in_corpus\endcsname{2}
\expandafter\def\csname paper@num@n@rel.secondary_host_records_in_corpus\endcsname{34}
\expandafter\def\csname paper@num@n@timing.cpu_count\endcsname{8}
\expandafter\def\csname paper@num@n@timing.processes\endcsname{4}
\expandafter\def\csname paper@num@n@timing.seconds\endcsname{812}
\expandafter\def\csname paper@num@pct0@audit.corpus.largest_share_of_judged_fields.share\endcsname{40\%}
\expandafter\def\csname paper@num@pct0@diag.suites.main35.overall.headroom.0.all_share\endcsname{44\%}
\expandafter\def\csname paper@num@pct0@facts.early_exit.calls_saved_share\endcsname{54\%}
\expandafter\def\csname paper@num@pct0@facts.early_exit.cost_saved_share\endcsname{43\%}
\expandafter\def\csname paper@num@pct0@facts.operating_points.cascade_gap_closed_by_lifting_the_cap\endcsname{97\%}
\expandafter\def\csname paper@num@pct0@facts.operating_points.confidence_cascade.uncapped.cost_share_of_best_single\endcsname{123\%}
\expandafter\def\csname paper@num@pct0@facts.operating_points.router_cost_relative_to_control.uncapped\endcsname{88\%}
\expandafter\def\csname paper@num@pct0@facts.operating_points.single_by_cost.capped.cost_share_of_best_single\endcsname{85\%}
\expandafter\def\csname paper@num@pct0@facts.relative_to_best_single.confidence_cascade.cost_ratio\endcsname{47\%}
\expandafter\def\csname paper@num@pct0@facts.relative_to_best_single.knn_router.cost_ratio\endcsname{63\%}
\expandafter\def\csname paper@num@pct0@fu.ridge.largest_cost_share\endcsname{72\%}
\expandafter\def\csname paper@num@pct0@fu.unknown.chance_share\endcsname{3\%}
\expandafter\def\csname paper@num@pct0@fu.unknown.ridge_cost_share\endcsname{64\%}
\expandafter\def\csname paper@num@pct0@fu.unknown.same_benchmark_share\endcsname{90\%}
\expandafter\def\csname paper@num@pct0@homog.suites.main35.all_models.none_right_unanimous\endcsname{42\%}
\expandafter\def\csname paper@num@pct0@homog.suites.main35.all_pairs.both_wrong_agreement\endcsname{72\%}
\expandafter\def\csname paper@num@pct0@homog.suites.main35.all_pairs.chance_agreement\endcsname{26\%}
\expandafter\def\csname paper@num@pct1@audit.vectors.off_sum_share\endcsname{2.9\%}
\expandafter\def\csname paper@num@pct1@bm.suites.main35.references.T.accuracy\endcsname{81.9\%}
\expandafter\def\csname paper@num@pct1@diag.suites.main35.overall.headroom.0.none_share\endcsname{4.0\%}
\expandafter\def\csname paper@num@pct1@facts.no_target_labels.transfer_single_accuracy\endcsname{79.9\%}
\expandafter\def\csname paper@num@pct1@fu.own.isarcasm_own_reference\endcsname{74.2\%}
\expandafter\def\csname paper@num@pct1@fu.own.isarcasm_stack\endcsname{26.0\%}
\expandafter\def\csname paper@num@pct1@fu.own.own_reference_own\endcsname{81.0\%}
\expandafter\def\csname paper@num@pct1@fu.own.reference_own\endcsname{78.7\%}
\expandafter\def\csname paper@num@pct1@fu.unknown.across_accuracy\endcsname{85.5\%}
\expandafter\def\csname paper@num@pct1@hard.suites.main35.few_right.accuracy.best_single\endcsname{31.1\%}
\expandafter\def\csname paper@num@pct1@hard.suites.main35.few_right.accuracy.stack_final_temperature\endcsname{26.4\%}
\expandafter\def\csname paper@num@pct1@hard.suites.main35.few_right.accuracy.uniform_mean\endcsname{1.0\%}
\expandafter\def\csname paper@num@pct1@hard.suites.main35.few_right.share\endcsname{5.6\%}
\expandafter\def\csname paper@num@pct1@rows.headline.best_single.accuracy\endcsname{86.3\%}
\expandafter\def\csname paper@num@pts1@contrasts.C1.headline.accuracy.difference\endcsname{2.7}
\expandafter\def\csname paper@num@pts1@contrasts.C5.headline.accuracy.difference\endcsname{1.0}
\expandafter\def\csname paper@num@pts1@contrasts.C9.headline.accuracy.difference\endcsname{3.3}
\expandafter\def\csname paper@num@pts1@contrasts.D1.headline.accuracy.difference\endcsname{0.7}
\expandafter\def\csname paper@num@pts2@contrasts.C1.headline.accuracy.ci_high\endcsname{2.38}
\expandafter\def\csname paper@num@pts2@contrasts.C1.headline.accuracy.ci_low\endcsname{3.04}
\expandafter\def\csname paper@num@pts2@contrasts.C5.headline.accuracy.ci_high\endcsname{1.12}
\expandafter\def\csname paper@num@pts2@contrasts.C5.headline.accuracy.ci_low\endcsname{0.79}
\expandafter\def\csname paper@num@pts2@contrasts.C8.headline.accuracy.ci_high\endcsname{0.58}
\expandafter\def\csname paper@num@pts2@contrasts.C8.headline.accuracy.ci_low\endcsname{1.05}
\expandafter\def\csname paper@num@pts2@contrasts.C8.headline.accuracy.difference\endcsname{0.81}
\expandafter\def\csname paper@num@pts2@fu.control.ridge_minus_control.difference\endcsname{0.08}
\expandafter\def\csname paper@num@pts2@fu.folds.C5_frozen\endcsname{0.96}
\expandafter\def\csname paper@num@pts2@fu.folds.C5_mean\endcsname{1.01}
\expandafter\def\csname paper@num@pts2@fu.folds.C8_frozen\endcsname{0.81}
\expandafter\def\csname paper@num@pts2@fu.folds.C8_mean\endcsname{0.67}
\expandafter\def\csname paper@num@pts2@fu.folds.D1_frozen\endcsname{0.72}
\expandafter\def\csname paper@num@pts2@fu.folds.D1_mean\endcsname{0.81}
\expandafter\def\csname paper@num@pts2@fu.full_text.knn_minus_best.difference\endcsname{0.61}
\expandafter\def\csname paper@num@pts2@fu.own.router_minus_best\endcsname{0.32}
\expandafter\def\csname paper@num@pts2@fu.own.router_minus_own\endcsname{2.00}
\expandafter\def\csname paper@num@pts2@fu.own.stack_minus_best\endcsname{1.00}
\expandafter\def\csname paper@num@pts2@fu.own.stack_minus_own\endcsname{1.32}
\expandafter\def\csname paper@num@pts2@fu.own.text_router_minus_own\endcsname{1.49}
\expandafter\def\csname paper@num@pts2@fu.pool.balanced_minus_uniform.difference\endcsname{0.45}
\expandafter\def\csname paper@num@pts2@fu.unknown.across_capped_minus_text.difference\endcsname{0.23}
\expandafter\def\csname paper@num@pts2@fu.unknown.across_minus_best.difference\endcsname{0.80}
\expandafter\def\csname paper@num@pts2@fu.unknown.across_minus_global.difference\endcsname{4.05}
\expandafter\def\csname paper@num@pts2@fu.unknown.ridge_minus_text.difference\endcsname{0.53}
\expandafter\def\csname paper@num@pts2@fu.unknown.ridge_uncapped_minus_best.difference\endcsname{0.18}
\expandafter\def\csname paper@num@pts2@ops.suites.main35.contrasts.mean_cascade_capped_vs_single.accuracy.difference\endcsname{2.91}
\expandafter\def\csname paper@num@pts2@ops.suites.main35.contrasts.mean_cascade_uncapped_vs_single.accuracy.difference\endcsname{0.14}
\expandafter\def\csname paper@num@s3@ops.suites.main35.contrasts.router_uncapped_vs_control.log_loss.ci_high\endcsname{\ensuremath{+0.014}}
\expandafter\def\csname paper@num@s3@ops.suites.main35.contrasts.router_uncapped_vs_control.log_loss.ci_low\endcsname{\ensuremath{+0.006}}
\expandafter\def\csname paper@num@s4@fu.full_text.ridge_minus_best_log_loss.over_high\endcsname{\ensuremath{-0.0005}}
\expandafter\def\csname paper@num@s4@fu.full_text.ridge_minus_best_log_loss.over_low\endcsname{\ensuremath{-0.0121}}
\expandafter\def\csname paper@num@s4@fu.full_text.ridge_minus_best_log_loss.within_high\endcsname{\ensuremath{-0.0029}}
\expandafter\def\csname paper@num@s4@fu.full_text.ridge_minus_best_log_loss.within_low\endcsname{\ensuremath{-0.0086}}
\expandafter\def\csname paper@num@s4@ops.suites.main35.contrasts.cascade_uncapped_vs_single.log_loss.ci_high\endcsname{\ensuremath{+0.0051}}
\expandafter\def\csname paper@num@s4@ops.suites.main35.contrasts.cascade_uncapped_vs_single.log_loss.ci_low\endcsname{\ensuremath{-0.0006}}
\expandafter\def\csname paper@num@s4@ops.suites.main35.contrasts.cascade_uncapped_vs_single.log_loss.difference\endcsname{\ensuremath{+0.0022}}
\expandafter\def\csname paper@num@s4@ops.suites.main35.contrasts.router_uncapped_vs_single.log_loss.ci_high\endcsname{\ensuremath{+0.0139}}
\expandafter\def\csname paper@num@s4@ops.suites.main35.contrasts.router_uncapped_vs_single.log_loss.ci_low\endcsname{\ensuremath{+0.0061}}
\expandafter\def\csname paper@num@s4@ops.suites.main35.contrasts.router_uncapped_vs_single.log_loss.difference\endcsname{\ensuremath{+0.0099}}
\expandafter\def\csname paper@num@s4@ops.suites.main35.contrasts.text_router_uncapped_vs_single.log_loss.ci_high\endcsname{\ensuremath{+0.0077}}
\expandafter\def\csname paper@num@s4@ops.suites.main35.contrasts.text_router_uncapped_vs_single.log_loss.ci_low\endcsname{\ensuremath{-0.0013}}
\expandafter\def\csname paper@num@s4@ops.suites.main35.router_by_features.at_most_one.capped_log_loss_difference\endcsname{\ensuremath{+0.0250}}
\expandafter\def\csname paper@num@s4@ops.suites.main35.router_by_features.two_or_more.capped_log_loss_difference\endcsname{\ensuremath{-0.0015}}
\expandafter\def\csname paper@num@spts2@fu.control.ridge_minus_control.over_high\endcsname{\ensuremath{+0.32}}
\expandafter\def\csname paper@num@spts2@fu.control.ridge_minus_control.over_low\endcsname{\ensuremath{-0.16}}
\expandafter\def\csname paper@num@spts2@fu.full_text.knn_minus_stored.difference\endcsname{\ensuremath{-0.04}}
\expandafter\def\csname paper@num@spts2@fu.full_text.knn_minus_stored.over_high\endcsname{\ensuremath{+0.14}}
\expandafter\def\csname paper@num@spts2@fu.full_text.knn_minus_stored.over_low\endcsname{\ensuremath{-0.22}}
\expandafter\def\csname paper@num@spts2@fu.full_text.knn_minus_stored.within_high\endcsname{\ensuremath{+0.17}}
\expandafter\def\csname paper@num@spts2@fu.full_text.knn_minus_stored.within_low\endcsname{\ensuremath{-0.25}}
\expandafter\def\csname paper@num@spts2@fu.own.router_groups_high\endcsname{\ensuremath{+0.68}}
\expandafter\def\csname paper@num@spts2@fu.own.router_groups_low\endcsname{\ensuremath{-0.03}}
\expandafter\def\csname paper@num@spts2@fu.own.router_over_high\endcsname{\ensuremath{+2.25}}
\expandafter\def\csname paper@num@spts2@fu.own.router_over_low\endcsname{\ensuremath{-0.83}}
\expandafter\def\csname paper@num@spts2@fu.own.stack_groups_high\endcsname{\ensuremath{+1.30}}
\expandafter\def\csname paper@num@spts2@fu.own.stack_groups_low\endcsname{\ensuremath{+0.71}}
\expandafter\def\csname paper@num@spts2@fu.own.stack_over_high\endcsname{\ensuremath{+1.76}}
\expandafter\def\csname paper@num@spts2@fu.own.stack_over_low\endcsname{\ensuremath{+0.12}}
\expandafter\def\csname paper@num@spts2@fu.pool.top_k_minus_stacking.difference\endcsname{\ensuremath{-0.05}}
\expandafter\def\csname paper@num@spts2@fu.pool.top_k_minus_stacking.over_high\endcsname{\ensuremath{+0.13}}
\expandafter\def\csname paper@num@spts2@fu.pool.top_k_minus_stacking.over_low\endcsname{\ensuremath{-0.19}}
\expandafter\def\csname paper@num@spts2@fu.unknown.across_capped_minus_text.difference\endcsname{\ensuremath{-0.23}}
\expandafter\def\csname paper@num@spts2@fu.unknown.across_capped_minus_text.over_high\endcsname{\ensuremath{+0.40}}
\expandafter\def\csname paper@num@spts2@fu.unknown.across_capped_minus_text.over_low\endcsname{\ensuremath{-0.90}}
\expandafter\def\csname paper@num@spts2@fu.unknown.across_capped_minus_text.within_high\endcsname{\ensuremath{+0.05}}
\expandafter\def\csname paper@num@spts2@fu.unknown.across_capped_minus_text.within_low\endcsname{\ensuremath{-0.47}}
\expandafter\def\csname paper@num@spts2@fu.unknown.across_minus_global.over_high\endcsname{\ensuremath{+6.23}}
\expandafter\def\csname paper@num@spts2@fu.unknown.across_minus_global.over_low\endcsname{\ensuremath{+2.16}}
\expandafter\def\csname paper@num@spts2@fu.unknown.ridge_minus_best.difference\endcsname{\ensuremath{-0.04}}
\expandafter\def\csname paper@num@spts2@fu.unknown.ridge_minus_best.over_high\endcsname{\ensuremath{+0.24}}
\expandafter\def\csname paper@num@spts2@fu.unknown.ridge_minus_best.over_low\endcsname{\ensuremath{-0.36}}
\expandafter\def\csname paper@num@spts2@fu.unknown.ridge_minus_best.within_high\endcsname{\ensuremath{+0.11}}
\expandafter\def\csname paper@num@spts2@fu.unknown.ridge_minus_best.within_low\endcsname{\ensuremath{-0.20}}
\expandafter\def\csname paper@num@spts2@fu.unknown.ridge_uncapped_minus_best.over_high\endcsname{\ensuremath{+0.64}}
\expandafter\def\csname paper@num@spts2@fu.unknown.ridge_uncapped_minus_best.over_low\endcsname{\ensuremath{-0.17}}
\expandafter\def\csname paper@num@spts2@fu.unknown.ridge_uncapped_minus_best.within_high\endcsname{\ensuremath{+0.34}}
\expandafter\def\csname paper@num@spts2@fu.unknown.ridge_uncapped_minus_best.within_low\endcsname{\ensuremath{+0.01}}
\expandafter\def\csname paper@num@spts2@ops.suites.main35.contrasts.cascade_uncapped_vs_single.accuracy.ci_high\endcsname{\ensuremath{+0.03}}
\expandafter\def\csname paper@num@spts2@ops.suites.main35.contrasts.cascade_uncapped_vs_single.accuracy.ci_low\endcsname{\ensuremath{-0.22}}
\expandafter\def\csname paper@num@spts2@ops.suites.main35.contrasts.cascade_uncapped_vs_single.accuracy.difference\endcsname{\ensuremath{-0.09}}
\expandafter\def\csname paper@num@text@audit.corpus.first_listing\endcsname{2026-09-18}
\expandafter\def\csname paper@num@text@audit.corpus.last_listing\endcsname{2026-11-17}
\expandafter\def\csname paper@num@text@timing.numpy\endcsname{2.3.5}
\expandafter\def\csname paper@num@text@timing.pandas\endcsname{2.3.3}
\expandafter\def\csname paper@num@text@timing.pyarrow\endcsname{22.0.0}
\expandafter\def\csname paper@num@text@timing.python\endcsname{3.13.9}
\expandafter\def\csname paper@num@text@timing.scikit_learn\endcsname{1.7.2}
\expandafter\def\csname paper@num@text@timing.scipy\endcsname{1.16.3}
\expandafter\def\csname paper@num@usd2@audit.telemetry.cost_usd\endcsname{\$86.05}
\newcommand{\assetcaption}[1]{\csname paper@caption@#1\endcsname}
\expandafter\def\csname paper@caption@accuracy_cost\endcsname{Headline suite (35 benchmarks). Mean accuracy and cost by model index for the suite and each task area. Panel counts give contributing benchmarks. Counts that vary by model appear in parentheses, with unequal cost and accuracy counts separated by a slash. The cost axis is logarithmic.}
\expandafter\def\csname paper@caption@bubbles\endcsname{Accuracy against cost of each of the models (benchmarks weighted equally); the area of a bubble is proportional to the model's median latency per request, averaged over benchmarks.}
\expandafter\def\csname paper@caption@donuts\endcsname{Share of requests sent to each model by the kNN router under the cost cap and without it, and the model whose answer the confidence cascade returns. Models below 5\% are grouped as Others.}
\expandafter\def\csname paper@caption@frontier_cost\endcsname{Headline suite (35 benchmarks). Accuracy of the evaluated operating points. The step line connects nondominated points in increasing cost. Each point includes 35 benchmarks.}
\expandafter\def\csname paper@caption@frontier_latency\endcsname{Headline suite (35 benchmarks). Accuracy of the evaluated operating points. The step line connects nondominated points in increasing latency. Each point includes 35 benchmarks.}
\expandafter\def\csname paper@caption@gains\endcsname{Best achievable accuracy gain (PerfGain) and largest cost saving at no loss of accuracy (CostSave) of each policy over its settings, against Best Single $\mathcal{B}$ (one model for all benchmarks, chosen on the test folds, as in LLMRouterBench) and against the best model per benchmark $\mathcal{B}_d$. Settings are chosen on the test folds. N/A: no setting is as accurate as the reference; arrows: the cheapest such setting costs more than twice the reference.}
\expandafter\def\csname paper@caption@hardness\endcsname{Top: judged questions by how many of the models are right on them. Bottom: accuracy of the shown methods on those questions. Benchmarks weighted equally.}
\expandafter\def\csname paper@caption@headroom\endcsname{Headroom by benchmark. Left: the share of judged questions that the most accurate single model gets right (dot) and that at least one model gets right (tick). Other columns: the share that at least two models, a majority, all models and no model get right.}
\expandafter\def\csname paper@caption@label_budget\endcsname{Log loss (left) and accuracy (right) on the headline suite of the best model per benchmark and the stacked panel, fitted on a sample of labelled source groups of the target benchmark (mean of 20 draws). Dashed: a model chosen on the other task areas with no target label. With all groups both use the four folds that are not the test fold.}
\expandafter\def\csname paper@caption@metric_bars\endcsname{Log loss, average accuracy and accuracy difference to the best model per benchmark ($\mathcal{B}_d$). The accuracy axis starts at 80\%.}
\expandafter\def\csname paper@caption@operating_points\endcsname{Headline suite (35 benchmarks). Accuracy against cost (left) and against mean latency (right). Each line is one method over its settings; black symbols are single configurations; grey dots are the single models. 1 single model lies below the plotted range.}
\expandafter\def\csname paper@caption@panel_size\endcsname{Log loss on the headline suite by panel size: equal-weight pools of the $k$ best-ranked models (top-k pool) and of $k$ random models (random pool, mean over seeds), and greedy selection with $k$ members.}
\expandafter\def\csname paper@caption@pareto\endcsname{Average accuracy against cost for the single models and every setting of the adaptive policies, with the empirical Pareto frontier. Control: the per-benchmark control, Black symbols: $\mathcal{B}_d$, stacking, early exit and Best Single $\mathcal{B}$.}
\expandafter\def\csname paper@caption@reliability\endcsname{Headline suite (35 benchmarks). Reliability by model index. The dashed line indicates equality. Counts under each panel give contributing benchmarks in increasing bin order, read across each line. Empty bins interrupt the curve.}
\expandafter\def\csname paper@caption@similarity\endcsname{Mean correlation between the errors of two models on the headline suite, for all pairs of the 17 models. Models are grouped by provider (white lines); the diagonal is left blank. 21 pairs are of two models of one provider.}

\newcommand{\literal}[1]{#1}
\newcommand{\name}{DecisionReplay}
\definecolor{bestcell}{HTML}{9EC5F4}
\definecolor{secondcell}{HTML}{DCEAFB}
\definecolor{markyes}{HTML}{0CA30C}
\definecolor{markno}{HTML}{D03B3B}
\newcommand{\cmark}{\textcolor{markyes}{\ding{51}}}
\newcommand{\xmark}{\textcolor{markno}{\ding{55}}}
\newcommand{\pmark}{\textcolor{markyes}{(\ding{51})}}
\newcommand{\nmark}{--}
\newcommand{\finding}[1]{\paragraph{#1}}

\newcommand{\releasestatement}{We release the simulator, the corpus builder, all analysis code and the whole
corpus: for every model and request, the probability vectors with the token counts, the billed cost and the measured
latency (Appendix~\ref{sec:rights}). Every result of this paper can be recomputed from these files and the
benchmarks' own releases.}
\ifpapersynthetic
  \AddToHook{shipout/foreground}{%
    \put(.5\paperwidth,-1.5cm){\makebox[0pt]{%
      \color[gray]{.62}\fontsize{24}{26}\selectfont\bfseries MADE-UP DATA}}}
\fi
\title{\name: A Corpus and Offline Simulator for Ensembling Decision Models}
\author{Dean Ariel \quad Seffi Cohen \\
  The Stein Faculty of Computer and Information Science \\
  Ben-Gurion University of the Negev \\
  Be'er Sheva, Israel}
\begin{document}
\maketitle
\begin{abstract}
Comparing ensembles, routers and cascades requires the output of every model on every item, and its cost. Stored collections built for routing keep no per-option probabilities. We provide a dataset on which ensemble methods, routers and cascades are simulated quickly, without new model calls. DecisionReplay holds the probability vectors of 17 commercial decision models, endpoints that return a probability for every answer option, on 35 benchmarks, with the token counts, billed cost and measured latency of every call, all released. A deterministic simulator replays any policy that selects, combines or cascades them. One well-chosen model per benchmark is hard to beat: pooling all models loses 2.7 accuracy points to it, and stacking gains 1.0 points at 3.4 times its cost. The models are wrong together: two that are both wrong choose the same wrong option in 72\% of cases. The best router matches that model at 64\% of its cost, and certified early exit cuts the cost of a stacked ensemble by 43\% without changing an answer. DecisionReplay makes it cheap and fast to test whether a new method changes these results.
\end{abstract}
\section{Introduction}
\label{sec:introduction}

\begin{figure}[t]
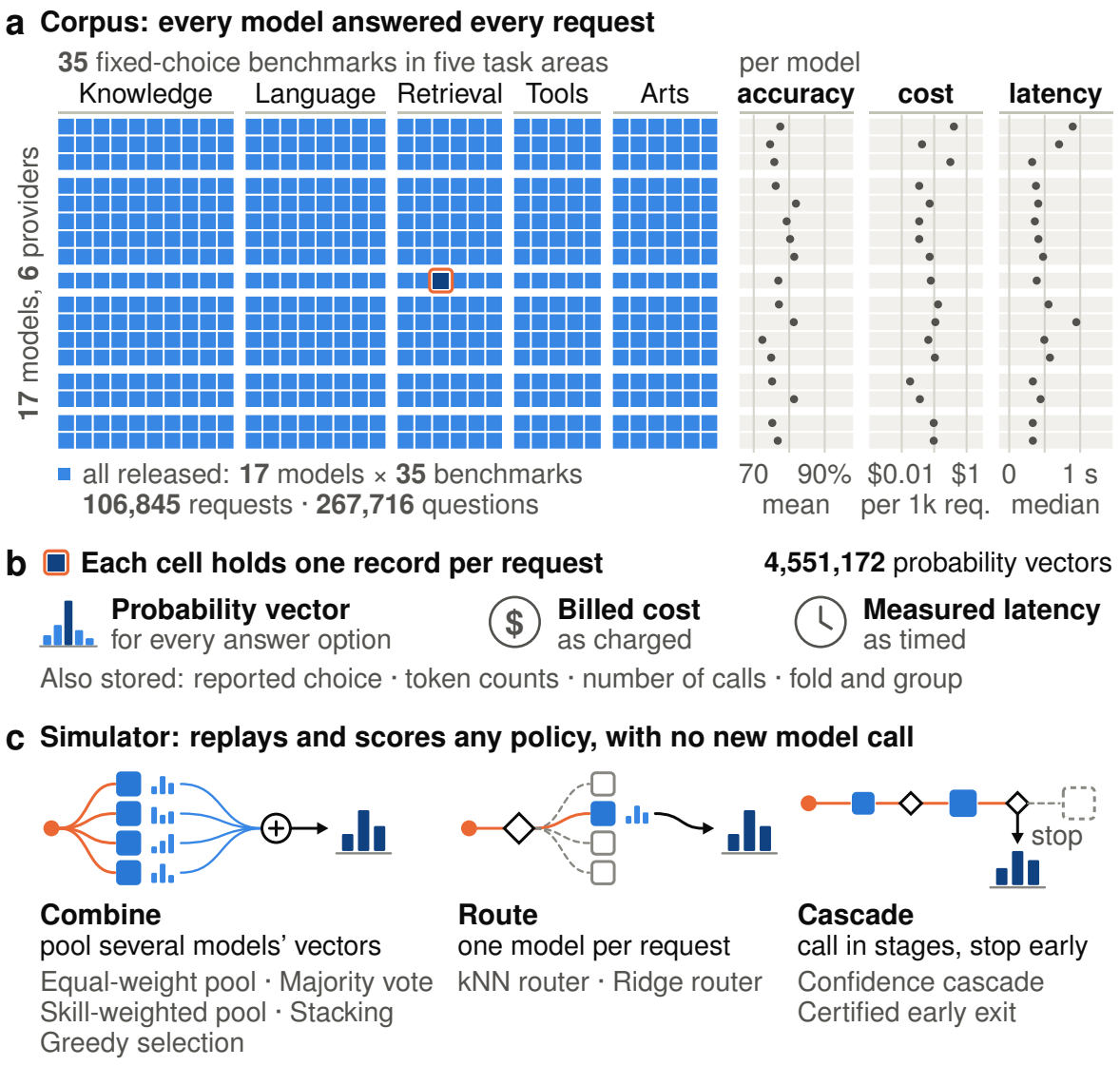

\centering
\definecolor{drink}{HTML}{0B0B0B}%
\definecolor{drgrey}{HTML}{52514E}%
\definecolor{drmid}{HTML}{898781}%
\definecolor{drstrip}{HTML}{F1F0EC}%
\definecolor{drline}{HTML}{D5D4CC}%
\definecolor{drrule}{HTML}{C3C2B7}%
\definecolor{drcell}{HTML}{3987E5}%
\definecolor{drmodel}{HTML}{2A78D6}%
\definecolor{drnavy}{HTML}{104281}%
\definecolor{drorange}{HTML}{EB6834}%
\definecolor{drtile}{HTML}{F3F3F1}%
\definecolor{drpale}{HTML}{B7D3F6}%
%
\caption{\name{} at a glance. (a)~The corpus: \num[n]{audit.corpus.models} decision models (rows, grouped by
provider) answered every request of \num[n]{audit.corpus.benchmarks} benchmarks (columns, grouped by task
area); dots give each model's mean accuracy, cost per \literal{1,000} requests (log scale) and median latency.
(b)~What each record holds. (c)~The three families of policies that the simulator replays.}
\label{fig:overview}
\end{figure}

Many model endpoints now answer the same request, at prices that differ by more than an order of magnitude and
with strengths that differ from task to task. Three families of methods try to exploit this:
\emph{ensembles} combine the outputs of several models, \emph{routers} send each request to the model expected to
do best, and \emph{cascades} call a cheap model first and a stronger one only when needed
\citep{yao2018using,ding2024hybrid,chen2023frugalgpt}.

Comparing these families fairly is expensive, because it needs the output of every model on every item.
Collections of stored outputs such as RouterBench and LLMRouterBench remove that cost
\citep{hu2024routerbench,li2026llmrouterbench}, but they were built for routing: they keep generated text with a
correctness score (Table~\ref{tab:comparison-checks}). Pooling, calibration and confidence-based stopping need a
probability for every answer option, which none of the routing collections reports storing, and the only one that
reports a latency per model estimates it from token counts.

We introduce \name{} (Figure~\ref{fig:overview}). It holds the outputs of \num[n]{audit.corpus.models} commercial
\emph{decision models}, endpoints that return a probability for every answer option in one call, on
\num[n]{audit.corpus.benchmarks} benchmarks from five task areas. For every model it stores the
probability vectors of \num[n]{audit.corpus.fields} questions, asked in \num[n]{audit.corpus.requests} requests, with the token counts, the billed cost and the measured latency of every call, and all of it is released at 
 \url{https://github.com/Seffi-Robust-AI-Lab/DecisionReplay}. \\
A deterministic simulator replays from these records any policy that selects (performance-oriented or cost-oriented), combines or cascades models, adds up its
recorded cost and latency, and scores it by accuracy and log loss.

With \name{}, we evaluate ensemble methods across the three families using a single protocol. The best model per benchmark,
chosen without the test folds, reaches \num[pct1]{rows.headline.best_single.accuracy} accuracy and proves hard to
beat. Pooling all models loses \num[pts1]{contrasts.C1.headline.accuracy.difference} points to it, and stacking gains
\num[pts1]{contrasts.C5.headline.accuracy.difference} points at
\num[f1]{facts.relative_to_best_single.stack_final_temperature.cost_ratio} times its cost. The reason is that the
models are \emph{wrong together}: their errors correlate at
\num[f2]{diag.suites.main35.overall.across_models.pairs.0.error_correlation} on average, and two models that are both
wrong choose the same wrong option in \num[pct0]{homog.suites.main35.all_pairs.both_wrong_agreement} of cases. Whether
routing helps depends on the reference: a nearest-neighbour router is more accurate than Best Single, the reference
of LLMRouterBench, but less accurate than the best model per benchmark, and a router that is not told the benchmark
finds it from the request's embedding. A ridge router matches the best model per benchmark at
\num[pct0]{fu.unknown.ridge_cost_share} of its cost. Finally, certified early exit cuts the cost of
a stacked ensemble by \num[pct0]{facts.early_exit.cost_saved_share} without changing an answer, but not its
latency.

Our contributions are as follows:
\begin{itemize}
\item \textbf{A released corpus of probability-level outputs with cost and latency.} \num[n]{audit.corpus.vectors}
probability vectors of \num[n]{audit.corpus.models} commercial models on \num[n]{audit.corpus.benchmarks}
benchmarks, with the token counts, billed cost and measured latency of every call.
\item \textbf{An offline simulator with unified baselines and metrics.} One interface for combination, routing and cascade policies, replay of recorded cost and latency, a fixed five-fold protocol and paired bootstrap intervals. Researchers can use it to investigate new ensemble methods without new model calls.
\item \textbf{An evaluation of the three families under one protocol.} It shows why combining adds little, separates
what routing gains from what knowing the benchmark gives, and tests whether more models and richer representations
of the request help.
\end{itemize}
\section{Related Work}
\label{sec:related}

\begin{table*}[tp]
\centering\footnotesize\setlength{\tabcolsep}{3.6pt}
\begin{tabular}{@{}lccccccrrc@{}}
\toprule
 & \multicolumn{3}{c}{Evaluated} & \multicolumn{3}{c}{Stored per model and item} & & & \\
\cmidrule(lr){2-4}\cmidrule(lr){5-7}
Collection & Comb. & Rout. & Casc. & Probs. & Cost & Latency & Datasets & Models & Release \\
\midrule
RouterBench~\citep{hu2024routerbench} & \xmark & \cmark & \cmark & \xmark & price list & \xmark & 8 & 11 & \cmark \\
EmbedLLM~\citep{zhuang2024embedllm} & \xmark & \cmark & \xmark & \xmark & \xmark & \xmark & 10 & 112 & \cmark \\
RouterEval~\citep{huang2025routereval} & \xmark & \cmark & \xmark & \nmark & \xmark & \nmark & 12 & 8,576 & \cmark \\
FusionFactory~\citep{feng2026fusionfactory} & \pmark & \cmark & \xmark & \xmark & price list & \xmark & 14 & 20 & \cmark \\
RouterArena~\citep{lu2026routerarena} & \xmark & \cmark & \xmark & \xmark & price list & \xmark & 23 & per router & \pmark\\
LLMRouterBench~\citep{li2026llmrouterbench} & \xmark & \cmark & \cmark & \xmark & billed or list & estimated & 21 & 33 & \cmark \\
DecisionPivot~\citep{zhuang2026decisionpivot} & \xmark & \xmark & \xmark & \cmark & tokens only & \xmark & 48 & 54 & \nmark \\
\midrule
\textbf{DecisionReplay} & \cmark & \cmark & \cmark & \cmark & billed & measured & 35 & 17 & \cmark \\
\bottomrule
\end{tabular}
\caption{Comparison with stored collections of model outputs. Comb., Rout., Casc.: combination, routing and cascade methods evaluated; Probs.: a probability for every answer option. \cmark{} yes, \xmark{} no, \pmark{} in part, \nmark{} not stated. Release: per-item outputs of every model are public.}
\label{tab:comparison-checks}
\end{table*}

\paragraph{Combining model outputs.} An equal-weight linear pool of perfectly calibrated members is miscalibrated as soon as two members differ~\citep{rahaman2020uncertainty}, and when point forecasts are combined, estimated weights often lose to the simple average~\citep{claeskens2016forecast}. Skills can be identified from agreement alone under the one-coin model~\citep{ma2019gradient}, but dependence among voters biases such estimates~\citep{jaffe2015unsupervised}. Language models are often wrong together: on leaderboard data two wrong models picked the same wrong option far more often than chance, more so within a company~\citep{kim2025correlated}, and in controlled experiments training on the same data reliably made outcomes more homogeneous~\citep{bommasani2022picking}. Choosing members for diversity as well as competence can help: DFPE clusters language models by embeddings of their responses, filters them by competence and weights them by their performance on the subject, and reports a gain of \literal{17.1} accuracy points over the best single model on MMLU-Pro~\citep{cohen2026dfpe}. Choosing the best of several noisy estimates also biases the winner upward~\citep{hu2026winner}, and in one study the best single agent chosen on twenty labelled questions beat every aggregator tested~\citep{bertiquille2026where}. When a language model aggregates generated responses, samples from the best single model can beat a mixture of different models~\citep{li2025rethinkingmoa}.

\paragraph{Routing and cascades.} Routers choose one model per query, from correctness predictors over embedded inputs~\citep{shnitzer2023large}, an encoder that predicts query difficulty~\citep{ding2024hybrid} or a factorised correctness matrix~\citep{zhuang2024embedllm}, or are trained on preference data~\citep{ong2024routellm}. Cascades stop on a learned score of the answer~\citep{chen2023frugalgpt}, on the small model's check of its own answer~\citep{aggarwal2023automix} or on the current model's confidence~\citep{jitkrittum2023when}; cascade routing routes, at each step, among all sequences of models that extend the ones already run~\citep{dekoninck2024unified}. Simple references are hard to beat: outside its RAG data, neither predictive router of RouterBench significantly beat its zero router~\citep{hu2024routerbench}, and in LLMRouterBench several routers did not beat the best single model~\citep{li2026llmrouterbench}. Decision models, which answer typed questions with a probability per option in one pass, have been studied as single models~\citep{chen2026large,yan2026chinese}; the closest work on several of them pairs a fast model with a slow one that may correct it~\citep{lin2026streamdecisionbench}.

\paragraph{Benchmarks of stored outputs.} RouterBench, EmbedLLM, RouterEval, FusionFactory, RouterArena and LLMRouterBench store, for each model and query, generated text or a correctness label, with a cost computed from price lists where they record one~\citep{hu2024routerbench,zhuang2024embedllm,huang2025routereval,feng2026fusionfactory,lu2026routerarena,li2026llmrouterbench} (Table~\ref{tab:comparison-checks}). LLMRouterBench estimates latency from token counts and serving statistics~\citep{li2026llmrouterbench}. DecisionPivot stores a probability for every option for \literal{54} systems, \literal{40} of them decision models, on \literal{48} tasks, but records no cost or latency and evaluates no combination, routing or cascade~\citep{zhuang2026decisionpivot}. Among the collections in Table~\ref{tab:comparison-checks}, only DecisionReplay stores per-option probabilities with the billed cost and measured latency of every call, and only it is used to evaluate all three families.
\section{\name{}}
\label{sec:resource}

\name{} supports the two settings that studies of model pools consider: combining models for the best
performance, and trading performance for cost.

\subsection{Model Pool}
\label{sec:pool}
A \emph{decision model} receives a state (text or structured data) and one or more typed questions, and returns
for each question a probability for every answer option, in a single call. We call one call a \emph{request} and
score its \emph{questions}. The pool has \num[n]{audit.corpus.models} commercial decision models from
\num[n]{audit.corpus.organisations} providers (Table~\ref{tab:model-pool}). All were called in autumn \literal{2026} with the same requests through the OpenRouter gateway, which first listed them
between \num[text]{audit.corpus.first_listing} and \num[text]{audit.corpus.last_listing}. The most expensive
model costs \num[f0]{facts.models.cost_ratio} times as much per request as the cheapest. No provider documents how
the probabilities are computed, so we treat every endpoint as a black box.

\subsection{Benchmarks}
\label{sec:benchmarks}
The \num[n]{audit.corpus.benchmarks} benchmarks were cast into typed questions by a public benchmark
kit~\citep{passos2026decision}: each task becomes one or more questions with a closed set of answer options. They
cover five task areas: \textbf{Knowledge} (e.g.,
GPQA~\citep{rein2024gpqa}, MMLU-Pro~\citep{wang2024mmlu}, ChessBench~\citep{ruoss2024amortized}),
\textbf{Language} (e.g., ANLI~\citep{nie2020adversarial}, HellaSwag~\citep{zellers2019hellaswag},
ACOS~\citep{cai2021aspect}), \textbf{Retrieval}, which includes intent classification (e.g.,
\literal{BANKING77}~\citep{casanueva2020efficient},
BRIGHT~\citep{su2025bright}, HoVer~\citep{jiang2020hover}), \textbf{Tools} (e.g., BFCL~\citep{patil2025berkeley},
API-Bank~\citep{li2023apibank}) and \textbf{Arts} (e.g., \literal{POP909}~\citep{wang2020pop909}, New Yorker
captions~\citep{hessel2023androids}). A question is a yes/no question or a choice among
\num[n]{audit.corpus.options_min} to \num[n]{audit.corpus.options_max} options, and in
\num[n]{audit.corpus.structure.per_item_inventory} benchmarks the options change from item to item. We score the
\num[n]{audit.corpus.judged_fields} questions for which a benchmark gives an accepted answer, the \emph{judged}
questions (Table~\ref{tab:benchmarks}). We weight benchmarks equally, because one of them, ACOS, holds
\num[pct0]{audit.corpus.largest_share_of_judged_fields.share} of the judged questions. Two retrieval benchmarks,
ToolRet and BRIGHT, judge only the relevant candidates, so accuracy on them is recall. All results are means over
the \num[n]{audit.corpus.benchmarks} benchmarks, which we call the headline suite; Appendix~\ref{sec:full-results}
repeats them on three parts of it.

\begin{figure*}[t]
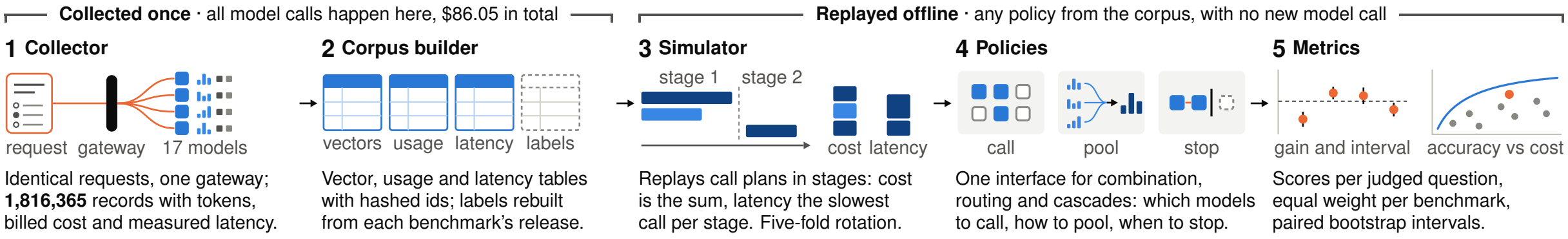

\centering
\definecolor{drink}{HTML}{0B0B0B}%
\definecolor{drgrey}{HTML}{52514E}%
\definecolor{drmid}{HTML}{898781}%
\definecolor{drstrip}{HTML}{F1F0EC}%
\definecolor{drline}{HTML}{D5D4CC}%
\definecolor{drrule}{HTML}{C3C2B7}%
\definecolor{drcell}{HTML}{3987E5}%
\definecolor{drmodel}{HTML}{2A78D6}%
\definecolor{drnavy}{HTML}{104281}%
\definecolor{drorange}{HTML}{EB6834}%
\definecolor{drtile}{HTML}{F3F3F1}%
\definecolor{drpale}{HTML}{B7D3F6}%
%
\caption{The modules of \name{}. The collector's records become the corpus; the simulator replays any policy from
it and the metrics score it, with no new model call.}
\label{fig:pipeline}
\end{figure*}

\subsection{Modular Design}
\label{sec:design}
\name{} has five modules (Figure~\ref{fig:pipeline}). The \emph{collector} sent each request to every model and
kept every call with its token counts, billed cost and measured latency. The \emph{corpus builder} turns these
records into tables of probability vectors, token usage and latency under hashed identifiers. Request text and gold
answers are not redistributed; a script rebuilds the labels from each benchmark's own release. The \emph{simulator} replays a \emph{policy}, which decides for each request which models to call, in
which stages, and how to turn their vectors into one vector per question. A request costs the sum of the billed
costs of its calls. Its latency is the sum over stages of the slowest call in each stage, since the calls of a stage
run in parallel. Each benchmark has five folds, and every fold is predicted by policies fitted on three of the others
and tuned on the fourth. \emph{Policies} share one interface, so a new method needs only a fitting rule and a
decision rule; Appendix~\ref{sec:baselines} gives those of our baselines. The \emph{metrics} module scores every
policy in the same way, and replay is deterministic (Appendix~\ref{sec:reproducibility}).

\subsection{Evaluation Metrics}
\label{sec:metrics}
\paragraph{Performance metrics.}
We report accuracy and log loss. Accuracy reads only the option that a method ranks first. Log loss is sensitive
to the probability itself: it falls as the accepted options receive more probability and rises steeply when a
method is confident in a wrong answer. For a method $a$ and a benchmark $d$ with judged questions $Q_d$,
\[
\mathrm{LogLoss}(a,d) = -\frac{\literal{1}}{|Q_d|} \sum_{q \in Q_d} \log \max(p_{a,q}, \varepsilon),
\]
where $p_{a,q}$ is the probability that $a$ gives to the accepted options of question $q$ and the floor
$\varepsilon$ is \literal{$10^{-15}$}. Both metrics average the benchmarks with equal weight. We compare each method with two
references. Best Single $\mathcal{B}$ is one model for all benchmarks, chosen on the test folds as in LLMRouterBench.
$\mathcal{B}_d$, the \emph{best model per benchmark}, is the model with the lowest log loss on the folds that are not
the test fold; it is our main reference. Where we compare two methods we give the difference in accuracy points.

\paragraph{Cost and latency.}
The cost of a policy is its billed cost per \literal{1,000} requests and its latency the mean replayed latency per
request, both averaged over benchmarks with equal weight. ParetoDist is the mean $L_1$ distance of a policy's
settings to the empirical Pareto frontier of accuracy and cost over all fitted policies and single models, on
min--max normalised axes with cost on a log scale.

\paragraph{Intervals.}
Differences come with two \literal{95}\% percentile intervals from \num[n]{facts.bootstrap.draws} paired bootstrap
draws. One resamples, within each benchmark, the source groups (questions that share a source document); the other
resamples benchmarks. The contrasts that we fixed in advance are adjusted with Holm's method
(Appendix~\ref{sec:setup}).

\begin{table}[tp]
\centering\footnotesize
\begin{tabular}{@{}lr@{}}
\toprule
Benchmarks & 35 \\
Task areas & 5 \\
Models & 17 \\
Providers & 6 \\
Requests & 106,845 \\
Questions & 267,716 \\
\quad with an accepted answer & 225,391 \\
Probability vectors & 4,551,172 \\
Model--request records & 1,816,365 \\
Input tokens & 1.55B \\
Collection cost & \$86.05 \\
\bottomrule
\end{tabular}
\caption{Overall statistics of DecisionReplay. Every model--request record holds the probability vectors of its questions, the token counts, the billed cost and the measured latency, and all of them are released; request text and gold answers are rebuilt from each benchmark's own release.}
\label{tab:statistics}
\end{table}

\subsection{Overall Statistics}
\label{sec:statistics}
Table~\ref{tab:statistics} summarises \name{}. After retries, every model returned a vector for every question (Appendix~\ref{sec:collection}), and collecting the corpus cost
\num[usd2]{audit.telemetry.cost_usd}. \releasestatement{}
\section{Experiments}
\label{sec:experiments}

\begin{figure*}[t]
\centering
\includegraphics{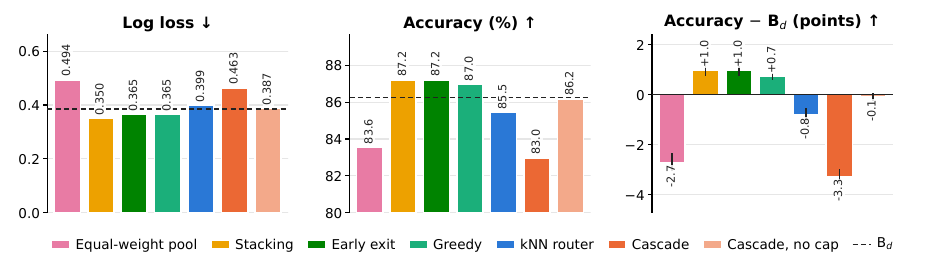}
\caption{\assetcaption{metric_bars}}
\label{fig:metric-bars}
\end{figure*}

\subsection{Experimental Setup}
\label{sec:experimental-setup}
We evaluate standard policies from the three families in two settings:
\begin{itemize}
\item \textbf{Performance-oriented setting}, where every model may be called: an equal-weight pool, the mean of all
models' vectors, and a majority vote; a pool weighted by skills estimated from agreement
alone~\citep{rozza2026jury}; the equal-weight pool with one temperature~\citep{rahaman2020uncertainty}; stacking of
calibrated members, with and without a final temperature~\citep{yao2018using}; and greedy forward
selection~\citep{caruana2004ensemble}.
\item \textbf{Performance--cost setting}: a $k$-nearest-neighbour router that sends a request to the model with the
best log loss on similar fitting requests, less a cost penalty; a control that picks one cost-penalised model per
benchmark; a ridge router that regresses each model's log score on the request's sentence embedding; a confidence
cascade over at most three models~\citep{jitkrittum2023when}; certified early exit, which calls the members of the
stacked panel in order of weight and stops once the uncalled weight cannot change an answer; and top-$k$ pools,
equal-weight pools of the $k$ best-ranked models.
\end{itemize}
We call the models that a policy combines its \emph{panel}. Every setting of a policy (panel size, cost penalty,
threshold) is tuned on the validation fold. For the router, its control and the cascade we take the best setting
whose validation cost does not exceed that of $\mathcal{B}_d$, and we show the cascade also without this
\emph{cost cap}. Unless stated, every policy is fitted inside each benchmark, so a router knows which benchmark a
request belongs to. Intervals in the text resample source groups within benchmarks;
Appendix~\ref{sec:full-results} also resamples benchmarks. Our earlier analyses of these outputs used the same folds,
so we read the intervals as descriptive, not as confirmatory tests. Appendices~\ref{sec:implementation}
and~\ref{sec:full-results} give the details.

\subsection{Results and Analyses}
\label{sec:results}

\begin{table*}[tp]
\centering\small\setlength{\tabcolsep}{5pt}
\begin{tabular}{@{}lrrrrrr@{}}
\toprule
Method & Log loss & Acc.\ (\%) & Own (\%) & Calls & USD/1k & Latency (s) \\
\midrule
\multicolumn{7}{@{}l}{\textit{Reference points}} \\
Best model per benchmark & 0.384 & 86.3 & 78.7 & 1.00 & 0.150 & 0.75 \\
Random model & 0.671 & 77.8 & 72.2 & 1.00 & 0.103 & 0.59 \\
\midrule
\multicolumn{7}{@{}l}{\textit{No labels}} \\
Equal-weight pool & 0.494 & 83.6 & 78.6 & 17.00 & 1.749 & 2.02 \\
Majority vote & 1.708 & 83.2 & 77.9 & 17.00 & 1.749 & 2.02 \\
Skill-weighted pool & 0.475 & 83.9 & 78.9 & 17.00 & 1.749 & 2.02 \\
\midrule
\multicolumn{7}{@{}l}{\textit{Supervised combination}} \\
Equal-weight pool + temperature & 0.426 & 83.6 & 78.6 & 17.00 & 1.749 & 2.02 \\
Stacking & 0.360 & 87.2 & 79.7 & 4.12 & 0.510 & 1.21 \\
Stacking + temperature & 0.350 & 87.2 & 79.7 & 4.12 & 0.510 & 1.21 \\
Greedy selection & 0.365 & 87.0 & 79.5 & 2.23 & 0.287 & 0.96 \\
\midrule
\multicolumn{7}{@{}l}{\textit{Routing and cascades}} \\
kNN router, request features & 0.399 & 85.5 & 79.0 & 1.00 & 0.094 & 0.66 \\
kNN router, text embeddings$^{\dagger}$ & 0.393 & 85.7 & 79.5 & 1.00 & 0.082 & 0.64 \\
Confidence cascade & 0.463 & 83.0 & 76.6 & 1.68 & 0.070 & 0.83 \\
\quad without the cost cap & 0.387 & 86.2 & 79.1 & 2.62 & 0.184 & 1.44 \\
Certified early exit & 0.365 & 87.2 & 79.7 & 1.89 & 0.293 & 1.32 \\
\bottomrule
\end{tabular}
\caption{Methods on the headline suite (35 benchmarks, equal weight). The own metric is each benchmark's native score. Latency is the mean per request, with the calls of one stage in parallel. The router and the cascade are shown at the chosen setting under a cost cap and without. $^{\dagger}$ needs the benchmark text.}
\label{tab:main-results}
\end{table*}

\subsubsection{Performance-Oriented Setting}

\begin{table}[tp]
\centering\scriptsize\setlength{\tabcolsep}{2.6pt}
\begin{tabular}{@{}llll@{}}
\toprule
Task area & First & Second & Third \\
\midrule
Knowledge & \textbf{Drex-Max-Fin 76.6} & Drex-Max 75.1 & D1-Ultra 74.9 \\
Language & \textbf{D1.2-Ultra 90.1} & D1-Ultra 89.5 & Decider 89.3 \\
Retrieval& \textbf{Drex-Max-Fin 81.4} & Decider 81.3 & D1-Ultra 80.8 \\
Tools & \textbf{Decider 94.4} & D1-Ultra 94.0 & D1.2-Ultra 93.8 \\
Arts & \textbf{D1.2-Ultra 75.8} & Drex-Max-Fin 75.4 & D1-Ultra 74.5 \\
\bottomrule
\end{tabular}
\caption{The three most accurate models of each task area, with their accuracy in percent}
\label{tab:top-models}
\end{table}

\finding{Pooling everything loses to the best model per benchmark; fitted weights win by little.}
The equal-weight pool of all models is \num[pts1]{contrasts.C1.headline.accuracy.difference} accuracy points below
$\mathcal{B}_d$, at
\num[f0]{facts.relative_to_best_single.uniform_mean.cost_ratio} times its cost (Figure~\ref{fig:metric-bars},
Table~\ref{tab:main-results}). A majority vote, skill weights and a temperature leave the pool below $\mathcal{B}_d$.
Only fitted weights overtake it: stacking with a final temperature is
\num[pts1]{contrasts.C5.headline.accuracy.difference} points ahead at
\num[f1]{facts.relative_to_best_single.stack_final_temperature.cost_ratio} times its cost, and greedy selection
\num[pts1]{contrasts.D1.headline.accuracy.difference} points ahead at
\num[f1]{facts.relative_to_best_single.greedy.cost_ratio} times. A top-$k$ pool with $k$ chosen on validation is
level with stacking in accuracy (Appendix~\ref{sec:further-analyses}). No single model wins everywhere:
\num[n]{facts.restyle.complementarity.benchmark_winners} of the \num[n]{facts.restyle.complementarity.models} models
are the most accurate on at least one benchmark (Table~\ref{tab:top-models}), which is why the choice per benchmark
matters.

\begin{figure}[t]
\centering
\includegraphics{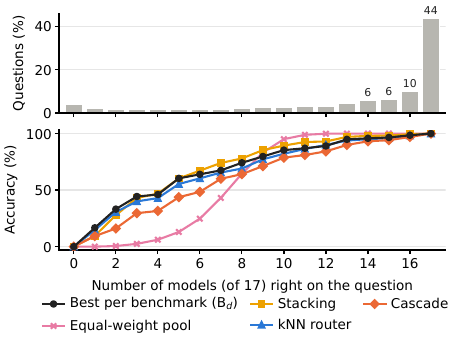}
\caption{\assetcaption{hardness}}
\label{fig:hardness}
\end{figure}

\finding{The models are wrong together.}
The errors of two models correlate at \num[f2]{diag.suites.main35.overall.across_models.pairs.0.error_correlation} on
average. Much of this is shared difficulty: among questions on which equally many of the other models err, the
correlation falls to \num[f2]{homog.suites.main35.all_pairs.error_correlation_given_others}
(\num[f2]{homog.suites.main35.same_organisation.error_correlation_given_others} for two models of one provider). But the
errors also point to the same answer: two models that are both wrong choose the same wrong option in
\num[pct0]{homog.suites.main35.all_pairs.both_wrong_agreement} of cases, against
\num[pct0]{homog.suites.main35.all_pairs.chance_agreement} under a uniform choice among the wrong options, and under
an equal-correlation approximation the pool is worth
\num[f1]{diag.suites.main35.overall.headroom.0.effective_voters} independent voters
(Appendix~\ref{sec:models-appendix}). All models are right on
\num[pct0]{diag.suites.main35.overall.headroom.0.all_share} of the judged questions and none on
\num[pct1]{diag.suites.main35.overall.headroom.0.none_share} (Figure~\ref{fig:hardness}, top).

\finding{Where few models are right, a pool follows the majority.}
Figure~\ref{fig:hardness} sorts the judged questions by the number of models that answer them correctly. On the
\num[pct1]{hard.suites.main35.few_right.share} that only one to three models get right, every method in the figure is
below $\mathcal{B}_d$, which reaches \num[pct1]{hard.suites.main35.few_right.accuracy.best_single}: stacking reaches
\num[pct1]{hard.suites.main35.few_right.accuracy.stack_final_temperature} and the equal-weight pool
\num[pct1]{hard.suites.main35.few_right.accuracy.uniform_mean}, because a pooled vector favours the answer that most
members give. There the right models are less confident than the wrong ones (mean probability
\num[f2]{fu.few_right.confidence_right} against \num[f2]{fu.few_right.confidence_wrong}), so confidence does not find
them. The bins use the test labels: they describe the questions and cannot select them.

\begin{figure}[t]
\centering
\includegraphics[width=\columnwidth]{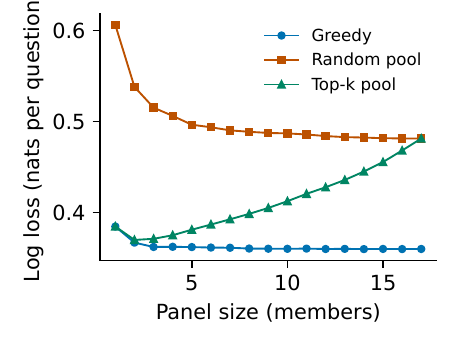}
\caption{\assetcaption{panel_size}}
\label{fig:panel-size}
\end{figure}

\finding{More models add little.}
An equal-weight pool of the best $k$ calibrated models is best with
\num[n]{facts.panel_size.ranked_top_k.best_size} members and worsens from there to the full panel, while greedy
selection is flat from a size of about \num[n]{facts.panel_size.greedy.first_size_within_0_001}
(Figure~\ref{fig:panel-size}). The strongest models are also the most alike: the errors of the five most accurate
ones correlate at \num[f2]{facts.restyle.top_models_error_correlation}, against
\num[f2]{facts.restyle.all_pairs_error_correlation} over all pairs.

\subsubsection{Performance--Cost Setting}

\begin{figure}[t]
\centering
\includegraphics{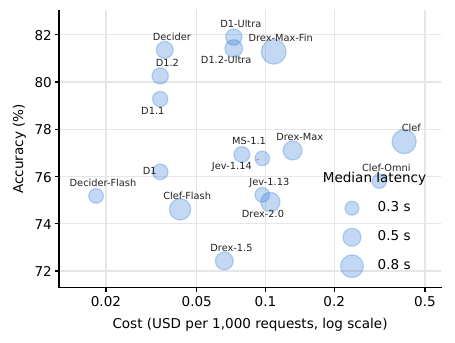}
\caption{\assetcaption{bubbles}}
\label{fig:bubbles}
\end{figure}

\finding{Cost and latency vary widely and are only weakly related to accuracy.}
Figure~\ref{fig:bubbles} places every model by accuracy, cost and median latency. Cost per request differs by a factor
of \num[f0]{facts.models.cost_ratio} between models, and neither cost nor latency is clearly related to accuracy
(Spearman \num[f2]{facts.models.spearman_accuracy_cost} and \num[f2]{facts.restyle.latency_accuracy_spearman}).
Latency is a third axis, which \name{} measures per call: two models that are within \literal{10}\% of each other
in cost and within one point in accuracy differ by a factor of \num[f1]{facts.restyle.latency.largest_ratio} in median
latency.

\finding{The best router matches one model per benchmark at lower cost.}
Best Single $\mathcal{B}$, the reference of LLMRouterBench, reaches
\num[pct1]{bm.suites.main35.references.T.accuracy}; $\mathcal{B}_d$ is
\num[f1]{facts.restyle.routing.reference_points_over_top} points more accurate at
\num[f1]{facts.restyle.routing.reference_cost_ratio_to_top} times the cost. The nearest-neighbour router lies between
them: it is \num[f1]{facts.restyle.routing.router_points_over_top} points above $\mathcal{B}$ and
\num[pts2]{contrasts.C8.headline.accuracy.difference} points below $\mathcal{B}_d$ (interval
\num[pts2]{contrasts.C8.headline.accuracy.ci_high} to \num[pts2]{contrasts.C8.headline.accuracy.ci_low}), at
\num[pct0]{facts.relative_to_best_single.knn_router.cost_ratio} of the cost of $\mathcal{B}_d$. A control that picks
one cost-penalised model per benchmark does better than the router: its log loss cannot be told from that of
$\mathcal{B}_d$, at \num[pct0]{facts.operating_points.single_by_cost.capped.cost_share_of_best_single} of the cost. A
router's gain over one global model can thus come from knowing the benchmark
(Appendix~\ref{sec:further-analyses}): a router that is not told the benchmark is as accurate as the text router
that is (\num[spts2]{fu.unknown.across_capped_minus_text.difference} points,
\num[spts2]{fu.unknown.across_capped_minus_text.within_low} to
\num[spts2]{fu.unknown.across_capped_minus_text.within_high}), and \num[pct0]{fu.unknown.same_benchmark_share} of its
neighbours come from the request's own benchmark. A ridge router on sentence embeddings, which we added after a first
draft, does best (Table~\ref{tab:representations}): it is as accurate as $\mathcal{B}_d$ (\num[spts2]{fu.unknown.ridge_minus_best.difference} points,
\num[spts2]{fu.unknown.ridge_minus_best.within_low} to \num[spts2]{fu.unknown.ridge_minus_best.within_high}) at
\num[pct0]{fu.unknown.ridge_cost_share} of its cost, and no variant of it is more accurate when benchmarks are
resampled (Appendix~\ref{sec:further-analyses}). When the task is known, a choice per benchmark is the reference to
beat, and a router earns its place by matching it for less.

\begin{table}[tp]
\centering\scriptsize\setlength{\tabcolsep}{2.8pt}
\begin{tabular}{@{}lrrr@{}}
\toprule
Representation & Log loss & Acc. (\%) & Cost \\
\midrule
Five request features & 0.399 & 85.5 & 63\% \\
\quad with all neighbours & 0.386 & 86.0 & 71\% \\
TF-IDF + SVD & 0.399 & 85.4 & 51\% \\
MiniLM-L6 encoder & 0.393 & 85.7 & 55\% \\
SmolLM2-360M states & 0.396 & 85.7 & 58\% \\
Whole request, MiniLM & 0.394 & 85.7 & 60\% \\
\midrule
Ridge router, MiniLM-L6 encoder & 0.380 & 86.2 & 64\% \\
\midrule
Per-benchmark control & 0.385 & 86.1 & 85\% \\
$\mathcal{B}_d$ & 0.384 & 86.3 & 100\% \\
\bottomrule
\end{tabular}
\caption{The kNN router with five representations of the request, under the cost cap; ``all neighbours'' adds every fitting request to the grid of neighbour counts. Cost: share of the best model per benchmark's cost. Text is cut at 256 tokens and leaves out the answer options, except for the whole request. The TF-IDF transform is fitted on the fitting folds only.}
\label{tab:representations}
\end{table}

\finding{Richer request representations barely help the router.}
Table~\ref{tab:representations} gives the router several representations of the request. The log losses of the
\num[n]{facts.restyle.representations.count} stored ones lie within
\num[f3]{facts.restyle.representations.log_loss_spread} nats of each other, all above that of $\mathcal{B}_d$, and
embedding the whole request, answer options included, leaves the nearest-neighbour router below $\mathcal{B}_d$
as well.

\finding{A cascade reaches the reference only when it may spend more.}
Under the cost cap the cascade stops early on most requests and is
\num[pts1]{contrasts.C9.headline.accuracy.difference} points below $\mathcal{B}_d$ at
\num[pct0]{facts.relative_to_best_single.confidence_cascade.cost_ratio} of its cost; the cap changed its threshold in
\num[n]{ops.suites.main35.selection.confidence_cascade.cap_changed_choice} of the
\num[n]{ops.suites.main35.selection.confidence_cascade.cells} combinations of benchmark and test fold. Without the
cap it calls \num[f1]{ops.suites.main35.points.confidence_cascade.uncapped.calls_per_request} models per request and
is level with $\mathcal{B}_d$ in accuracy
(\num[spts2]{ops.suites.main35.contrasts.cascade_uncapped_vs_single.accuracy.difference} points,
\num[spts2]{ops.suites.main35.contrasts.cascade_uncapped_vs_single.accuracy.ci_low} to
\num[spts2]{ops.suites.main35.contrasts.cascade_uncapped_vs_single.accuracy.ci_high}) at
\num[pct0]{facts.operating_points.confidence_cascade.uncapped.cost_share_of_best_single} of its cost
(Appendix~\ref{sec:operating-points}).

\begin{figure}[t]
\centering
\includegraphics{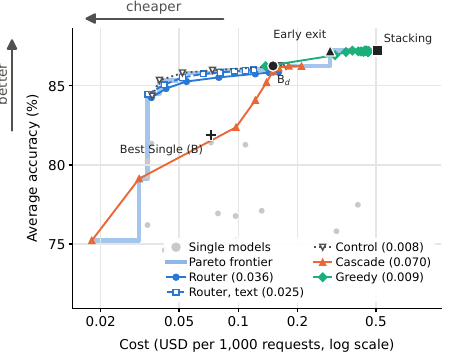}
\caption{\assetcaption{pareto}}
\label{fig:pareto}
\end{figure}

\finding{Certified early exit cuts the cost of a stacked panel without changing an answer, but not its latency.}
On the empirical Pareto frontier of accuracy against cost (Figure~\ref{fig:pareto}), the per-benchmark control, the
text router, greedy panels and early exit trace the upper part. Early exit returns exactly the answers of the stacked
panel with \num[pct0]{facts.early_exit.calls_saved_share} fewer calls and
\num[pct0]{facts.early_exit.cost_saved_share} less cost, at a slightly higher log loss. Because its calls are
sequential, its mean latency is higher than that of the stacked panel called in parallel
(\num[f2]{rows.headline.certified_exit.latency_mean_s}\,s against
\num[f2]{rows.headline.stack_final_temperature.latency_mean_s}\,s): early exit saves cost, not time.
\section{Conclusions}
\label{sec:conclusion}

We introduced \name{}, a released corpus of probability vectors with the token counts, billed cost and measured
latency of every call, and an offline simulator that replays combination, routing and cascade policies from it. One
well-chosen model per benchmark is hard to beat: pooling everything loses to it, fitted weights gain a point at more
than three times the cost, and the best router matches it at lower cost. The models are wrong together, which
limits what combining them can add. \name{} makes it cheap to test whether a new method changes these results.

\section*{Limitations}
\label{sec:limitations}

The pool is \num[n]{audit.corpus.models} endpoints of one product type on one gateway, collected once in autumn
\literal{2026}; prices, serving hosts and model versions change, and the training data of the models is
unknown and may include these benchmarks. Every benchmark was cast into questions with a closed set of options, so
the corpus says nothing about free text generation, and methods that need generated text, repeated samples or model
internals cannot be replayed on it.

\section*{Ethical Considerations}
\label{sec:ethics}

The corpus holds numeric outputs of commercial models under hashed identifiers. It contains no text written by or
about persons, and no annotators or participants were involved; all labels come from the benchmarks' own releases.
We do not redistribute benchmark text or gold answers. Several benchmarks restrict commercial use, redistribution
in plain text or use for training, and keeping their rows out of the corpus avoids adding a copy that could enter
training data.

Model outputs are subject to the terms of the model's provider, of the host that served it and of the gateway. We
reviewed these terms for every model.
The corpus holds only models for which the review flagged no clause that bars benchmarking or requires permission to
publish comparisons or to compile a dataset. The conditions that it found are attached to the released files as
notices, and releasing the usage and latency records is our own decision (Appendix~\ref{sec:rights}). Publishing comparisons of commercial models can affect their
providers: we report what was measured, with the collection period, and make no claim beyond these tasks. The
collection was small (\num[usd2]{audit.telemetry.cost_usd} in total), and replaying it needs no further model call.

\paragraph{Intended use.}
The corpus is meant for evaluating policies that select, combine or cascade the listed models on the listed tasks.
It is not meant for training or distilling other models: the terms that apply to several of the models restrict
such use, and
users who fit weights or routers on the outputs should check the notices of the models involved.

\clearpage
\appendix
\section{Benchmarks, Models and Release}
\label{sec:resources-appendix}

\begin{table*}[tp]
\centering\scriptsize\setlength{\tabcolsep}{2.8pt}
\begin{tabular}{@{}lllp{2.8cm}rrr@{}}
\toprule
Benchmark & Abbr. & Area & Own metric & Requests & Questions & Judged \\
\midrule
GPQA Diamond & GPQA & Knowledge & accuracy & 196 & 196 & 196 \\
WinoGrande & Wino & Knowledge & accuracy & 1,267 & 1,267 & 1,267 \\
GSM8K & GSM8K & Knowledge & mean accuracy over 4- and 10-choice tracks & 2,638 & 2,638 & 2,638 \\
ChessBench & Chess & Knowledge & best-move accuracy & 5,000 & 5,000 & 5,000 \\
MuSR & MuSR & Knowledge & accuracy & 752 & 752 & 752 \\
SATA-Bench & SATA & Knowledge & case exact accuracy & 1,650 & 15,517 & 15,517 \\
CRUXEval & CRUX & Knowledge & accuracy & 570 & 570 & 570 \\
CLadder & CLad & Knowledge & accuracy & 5,000 & 5,000 & 5,000 \\
MMLU-Pro & MMLUP & Knowledge & accuracy & 12,032 & 12,032 & 12,032 \\
BBH fixed-option tasks & BBH & Knowledge & accuracy & 5,507 & 5,507 & 5,507 \\
\addlinespace
ContractNLI & CNLI & Language & macro-F1 & 123 & 2,091 & 2,091 \\
ANLI & ANLI & Language & macro-F1 & 3,200 & 3,200 & 3,200 \\
HellaSwag & HSwag & Language & accuracy & 10,042 & 10,042 & 10,042 \\
ACOS & ACOS & Language & per-review F1 & 1,565 & 91,092 & 91,092 \\
FinEntity & FinE & Language & macro-F1 & 979 & 2,129 & 2,129 \\
iSarcasmEval & iSarc & Language & Sarcasm F1 · track A, English & 4,600 & 11,600 & 11,600 \\
VAST & VAST & Language & macro-F1 & 3,006 & 3,006 & 3,006 \\
NLI4CT & NLI4CT & Language & macro-F1 & 5,500 & 5,500 & 5,500 \\
\addlinespace
BANKING77 & BK77 & Retrieval & macro-F1 & 3,080 & 3,080 & 3,080 \\
CLINC150+OOS & CLINC & Retrieval & macro-F1 & 5,500 & 5,500 & 5,500 \\
BRIGHT & BRIGHT & Retrieval & nDCG@10 & 213 & 6,816 & 438 \\
Amazon ESCI & ESCI & Retrieval & macro-F1 & 5,000 & 5,000 & 5,000 \\
PhishNChips phishing decisions & Phish & Retrieval & accuracy & 2,000 & 18,000 & 2,000 \\
HoVer claim verification & HoVer & Retrieval & accuracy & 4,000 & 4,000 & 4,000 \\
\addlinespace
BFCL & BFCL & Tools & case exact accuracy & 1,694 & 4,768 & 4,768 \\
ToolRet & TRet & Tools & nDCG@10 & 651 & 20,831 & 884 \\
API-Bank & APIB & Tools & accuracy & 508 & 508 & 508 \\
Home appliance simulator & Home & Tools & case exact accuracy & 88 & 1,590 & 1,590 \\
When2Call MCQ & W2C & Tools & accuracy & 3,652 & 3,652 & 3,652 \\
\addlinespace
BPoMP & BPoMP & Arts & macro accuracy by variant & 5,000 & 5,000 & 5,000 \\
Humicroedit & Humi & Arts & accuracy & 2,628 & 2,628 & 2,628 \\
POP909-CL & POP & Arts & macro accuracy by song & 2,000 & 2,000 & 2,000 \\
cfcolor & cfcol & Arts & macro accuracy by user & 5,000 & 5,000 & 5,000 \\
Habermas Machine & Haber & Arts & group-preferred accuracy & 1,676 & 1,676 & 1,676 \\
New Yorker caption matching & NYC & Arts & accuracy & 528 & 528 & 528 \\
\bottomrule
\end{tabular}
\caption{The benchmarks with the abbreviations used in the tables, their task area, own metric and size.}
\label{tab:benchmarks}
\end{table*}
\begin{table*}[tp]
\centering\scriptsize\setlength{\tabcolsep}{3.4pt}
\begin{tabular}{@{}llrrll@{}}
\toprule
Model & Abbr. & Input price & Median & Documented & Review \\
 & & (\$ per 1M tokens) & latency (s) & base model & of terms \\
\midrule
\texttt{cloudflare/clef} & Clef & 0.192 & 0.54 & Qwen/\allowbreak{}Qwen3.8-27B & unverified \\
\texttt{cloudflare/clef-flash} & Clef-Flash & 0.019 & 0.50 & Qwen/\allowbreak{}Qwen3.5-9B & unverified \\
\texttt{cloudflare/clef-omni} & Clef-Omni & 0.150 & 0.25 & Qwen/\allowbreak{}Qwen3-Omni-30B-A3B-Instruct & cleared \\
\texttt{liquid/d1} & D1 & 0.020 & 0.32 & -- & conditional \\
\texttt{liquid/d1-ultra} & D1-Ultra & 0.042 & 0.34 & -- & conditional \\
\texttt{liquid/d1.1} & D1.1 & 0.020 & 0.31 & -- & conditional \\
\texttt{liquid/d1.2} & D1.2 & 0.020 & 0.35 & -- & conditional \\
\texttt{liquid/d1.2-ultra} & D1.2-Ultra & 0.042 & 0.38 & -- & conditional \\
\texttt{microsoft/microsoft-decision-1.1} & MS-1.1 & 0.042 & 0.35 & -- & cleared \\
\texttt{nace-ai/drex-max} & Drex-Max & 0.060 & 0.41 & -- & conditional \\
\texttt{nace-ai/drex-max-finance} & Drex-Max-Fin & 0.060 & 0.83 & -- & conditional \\
\texttt{nace-ai/drex-v1.5} & Drex-1.5 & 0.040 & 0.39 & XiaomiMiMo/\allowbreak{}MiMo-V2.6-Distill-Qwen-9B & conditional \\
\texttt{nace-ai/drex-v2.0} & Drex-2.0 & 0.060 & 0.45 & -- & conditional \\
\texttt{perplexity/pplx-decider-flash-v1.2} & Decider-Flash & 0.010 & 0.26 & Qwen/\allowbreak{}Qwen3.5-4B & unverified \\
\texttt{perplexity/pplx-decider-v1.2} & Decider & 0.020 & 0.35 & Qwen/\allowbreak{}Qwen3.8-27B & unverified \\
\texttt{typesafe/jev-1.13} & Jev-1.13 & 0.042 & 0.29 & -- & conditional \\
\texttt{typesafe/jev-1.14} & Jev-1.14 & 0.042 & 0.29 & -- & conditional \\
\bottomrule
\end{tabular}
\caption{The model pool: listed name, the abbreviation used in tables and figures, the listed input price (no model charged for output tokens), the median latency over all requests, the base model where the provider documents one, the status of its outputs in our review of terms.}
\label{tab:model-pool}
\end{table*}

\subsection{Benchmarks}
\label{sec:benchmark-sources}
Table~\ref{tab:benchmarks} lists the benchmarks with the abbreviations used in the tables. Their sources are
BFCL~\citep{patil2025berkeley}, ToolRet~\citep{shi2025retrieval}, API-Bank~\citep{li2023apibank},
\literal{BANKING77}~\citep{casanueva2020efficient}, \literal{CLINC150}~\citep{larson2019evaluation},
ContractNLI~\citep{koreeda2021contractnli}, ANLI~\citep{nie2020adversarial}, BPoMP~\citep{abdibayev2021bpomp},
Humicroedit~\citep{hossain2019president}, \literal{POP909}~\citep{wang2020pop909}, colour
preferences~\citep{odonovan2014collaborative}, GPQA~\citep{rein2024gpqa}, WinoGrande~\citep{sakaguchi2020winogrande},
HellaSwag~\citep{zellers2019hellaswag}, \literal{GSM8K}~\citep{cobbe2021training}, ChessBench~\citep{ruoss2024amortized},
MuSR~\citep{sprague2024musr}, SATA-Bench~\citep{xu2026sata}, BRIGHT~\citep{su2025bright}, Amazon
ESCI~\citep{reddy2022shopping}, ACOS~\citep{cai2021aspect}, FinEntity~\citep{tang2023finentity},
iSarcasmEval~\citep{abufarha2022semeval}, VAST~\citep{allaway2020zero}, \literal{NLI4CT}~\citep{jullien2024semeval},
CRUXEval~\citep{gu2024cruxeval}, CLadder~\citep{jin2023cladder}, the Habermas Machine
data~\citep{tessler2024ai}, PhishNChips~\citep{litvak2026system}, MMLU-Pro~\citep{wang2024mmlu},
BBH~\citep{suzgun2023challenging}, HoVer~\citep{jiang2020hover},
\literal{When2Call}~\citep{ross2025when2call} and the New Yorker caption data~\citep{hessel2023androids}; one
benchmark, a home-appliance simulator, is the kit's own~\citep{passos2026decision}. The kit turns each source into
typed questions with a closed set of options, so each benchmark here is a task derived from the original.

\subsection{Release and Terms}
\label{sec:rights}
Table~\ref{tab:model-pool} lists the models with the outcome of our review of terms. We reviewed the terms of the
gateway, of every provider and serving host, and of every benchmark source for one question: may the numeric outputs be redistributed
openly? The review classed a model as \emph{cleared} when the terms that we read allow it, \emph{conditional} when
use restrictions, a required permission or an open question of applicability qualify the answer, and
\emph{unverified} when the terms did not establish the scope of the grant on hosted outputs. It cleared
\num[n]{audit.rights.model_status_counts.cleared} models, found
\num[n]{audit.rights.model_status_counts.conditional} conditional and could not verify
\num[n]{audit.rights.model_status_counts.unverified}. All models were called under their standard terms.

The release is our decision, and it goes beyond what the review cleared in two respects. First, it includes the
conditional and unverified models. For none of them did the review flag a clause that bars benchmarking or requires
permission to publish comparisons or to compile a dataset; their conditions concern later use of the outputs, such
as developing other models, and are attached to the files as notices. Second, the review covered the probability
vectors and kept billing records and timing outside the public subset. We release, for every request and model, the
number of calls, the token counts, the billed cost and the measured latency, without the host that served the
request. One model was served on \num[n]{rel.secondary_host_cells_in_corpus} benchmarks partly by another host
(\num[n]{rel.secondary_host_records_in_corpus} requests). A further \num[n]{rel.excluded_requests} requests, in
\num[n]{rel.excluded_request_benchmarks} benchmarks, had been served for that model by a host whose terms condition
benchmarking of its service; we removed these requests from the corpus for every model.

We also collected outputs of \num[n]{rel.models_held} further models and on a further benchmark. The review flagged
clauses in the terms of these models that may bar benchmarking or require permission to publish comparisons or to
compile a dataset, and the source of the benchmark asks for a review before results are published. They were left
out for these terms and are part of neither this paper nor the release.

The source terms of the benchmarks are clear for \num[n]{audit.rights.benchmark_source_category_counts.clear}
benchmarks, restricted for \num[n]{audit.rights.benchmark_source_category_counts.restricted} and unclear for
\num[n]{audit.rights.benchmark_source_category_counts.unclear}. No benchmark text is redistributed, so the source
terms bear on the labels that a user rebuilds, not on the corpus files.
\section{Implementation Details}
\label{sec:implementation}

\subsection{Corpus}
\label{sec:collection}
\paragraph{Records.}
For every model and question the corpus stores the probability vector as returned, in option order, and the option
that the model reported as its choice. For every model and request it stores the number of calls, the input and
output token counts, the billed cost and the measured latency. The endpoints return no generated text,
log-probabilities or repeated samples, so the corpus holds none. Requests carry hashed identifiers, a fold and hashed
identifiers of their source groups. A script rebuilds the table of accepted options from benchmark rows that a user
obtains under each benchmark's own terms; the table rebuilt in this way is identical to ours.
\num[n]{leak.requests_with_duplicate_text_across_folds} requests in \num[n]{leak.benchmarks_affected} benchmarks
share their text with a request in another fold.

\paragraph{Collection.}
Collecting the corpus took \num[n]{audit.corpus.attempts} attempts, of which
\num[n]{audit.corpus.successful_physical_calls} calls succeeded, and cost \num[usd2]{audit.telemetry.cost_usd}.
The corpus has a vector for every model and question (\num[n]{audit.corpus.vectors} in all).
\num[n]{audit.corpus.adapted_observations} requests of
\num[n]{audit.corpus.models_answering_in_smaller_batches} models were rejected as a whole and answered in smaller
batches of questions; the record of such a request gives its number of calls. The billed
cost equals input tokens times the listed price on every request for
\num[n]{audit.telemetry.price_card_models_exact} of the models. For the remaining model it does so except on
\num[n]{audit.telemetry.price_card_rows_beyond_rounding} requests, which another host than its usual one served.
The number of attempts and the serving host come from our collection records; the released files do not hold them.

\paragraph{Decoding.}
Returned vectors are not always normalised: \num[pct1]{audit.vectors.off_sum_share} do not sum to one, and
\num[n]{audit.vectors.models_with_zero_entries} models return exact zeros, in
\num[n]{audit.vectors.zero_mass_on_accepted} cases on every accepted option, so log loss needs a floor. The
simulator renormalises each vector and decodes its first maximum. The model's reported choice differs from it in
\num[n]{audit.vectors.reported_differs_first} cases, all of them exact ties. A question can have several accepted options, and on a yes/no question the answer is ``yes'' when its probability is at least one half.

\subsection{Protocol}
\label{sec:setup}
\paragraph{Folds.}
Each benchmark has five fixed folds, and questions that share a source document or a scoring group lie in one fold.
Fold $f$ is predicted by policies that never saw it: a policy with a setting to tune is fitted on the three folds
other than $f$ and $f+\literal{1}$ and tuned on fold $f+\literal{1}$ (modulo five). Stacking and the pool's
temperature use three folds and tune nothing. $\mathcal{B}_d$ and the skill-weighted pool, which reads no labels, use
all four non-test folds.

\paragraph{Recalibration.}
Where a policy uses calibrated members, each model's vectors pass through one transform from a fixed menu (identity,
a temperature, or mixing with the uniform distribution), chosen per model and benchmark by log loss on the fitting
folds, estimated by cross-fitting.

\paragraph{Scores and intervals.}
Log loss clips the probability of the accepted options from below at \literal{$10^{-15}$}. This floor matters for
\num[n]{facts.floor.rows_moving_more_than_0_003} methods only. With a floor of \literal{$10^{-3}$}, the majority vote,
whose vote shares are often exactly zero, moves from \num[f3]{rows.headline.majority_vote.log_loss} to
\num[f3]{rows.headline.majority_vote.log_loss_floor_1e-3} nats and the random model from
\num[f3]{rows.headline.random_single.log_loss} to \num[f3]{rows.headline.random_single.log_loss_floor_1e-3}; no
other method of Table~\ref{tab:main-results} moves by more than \num[f3]{facts.floor.largest_shift_of_other_rows}
nats. Differences between policies
come with two percentile intervals from \num[n]{facts.bootstrap.draws} paired bootstrap draws, one resampling the
source groups within each benchmark and one resampling benchmarks. The planned contrasts of
Table~\ref{tab:contrasts} were fixed before the first run and are adjusted with Holm's method on log loss
(Appendix~\ref{sec:reproducibility} lists what changed since); all other comparisons are descriptive. The intervals
that resample benchmarks are
\num[f1]{facts.over_benchmarks_summary.narrowest_ratio} to \num[f0]{facts.over_benchmarks_summary.widest_ratio} times
wider than those that resample within benchmarks.

\paragraph{Cost cap.}
For the router, its control and the cascade, the cost weight $\lambda$ and the threshold $\tau$ are the grid values
with the lowest validation loss among those whose validation cost does not exceed that of $\mathcal{B}_d$, or the
cheapest value if none qualifies. Table~\ref{tab:settings} gives every grid, and
Appendix~\ref{sec:operating-points} reports the same policies without the cap.

\begin{table*}[tp]
\centering\footnotesize
\begin{tabular}{@{}llp{0.36\linewidth}@{}}
\toprule
Policy & Setting & Grid \\
\midrule
kNN router & neighbour count $k$ (tuned on the validation fold) & 5, 10, 20, 50 \\
kNN router & cost weight $\lambda$ & 0, 0.01, 0.02, 0.05, 0.1, 0.2, 0.5, 1 \\
Confidence cascade & threshold $\tau$; at most 3 models & 0, 0.5, 0.6, 0.7, 0.8, 0.9, 0.95, 0.99, 1.01 \\
Panels by size & number of members $k$ & 1 to 17 \\
Random panels & seeds & 1, 2, 3, 4, 5 \\
Label budgets & labelled source groups (20 draws each) & 4, 12, 48, all \\
Bootstrap & paired draws; seed & 2,000; 20261119 \\
\bottomrule
\end{tabular}
\caption{Grids and seeds of the experiments, from the settings file whose hash every result file records. The router and the cascade take the grid value chosen on the validation fold under the cost cap (Appendix~\ref{sec:setup}); the panels take a size chosen on the validation fold.}
\label{tab:settings}
\end{table*}

\subsection{Baselines}
\label{sec:baselines}
\paragraph{References.}
$\mathcal{B}_d$ is, for each test fold, the model with the lowest cross-fitted log loss on the four other folds,
recalibrated on them. Best Single $\mathcal{B}$ is the one model with the highest mean accuracy over the benchmarks;
it is chosen on the test folds, as in LLMRouterBench, and serves as a point of comparison only. The random model
$\mathcal{R}$ draws one model per request with a fixed seed.

\paragraph{Pools.}
The equal-weight pool is the mean of the vectors of all \num[n]{audit.corpus.models} models, and the majority vote
is the share of models that choose each option. The skill-weighted pool weights the models by reliabilities
estimated from agreement alone, with an expectation--maximisation fit of a one-coin model~\citep{rozza2026jury}; it
reads no labels. The pool with a temperature adds one temperature fitted on the fitting
folds~\citep{rahaman2020uncertainty}.

\paragraph{Stacking and selection.}
Stacking fits weights on the simplex over the calibrated members by log score on the fitting
folds~\citep{yao2018using}, with or without a final temperature on the pooled vector. Greedy selection is forward
selection with replacement from the calibrated members~\citep{caruana2004ensemble}, with the panel size chosen on the
validation fold.

\paragraph{kNN router and its control.}
The router is fitted inside each benchmark, so it knows which benchmark a request belongs to. It sends a request to
the model with the highest mean log score on the request's $k$ nearest fitting requests, minus a cost weight
$\lambda$ times the model's relative cost; $k$ and $\lambda$ are chosen on the validation fold. With every fitting
request as a neighbour, the router chooses one cost-penalised model per benchmark and test fold. This is its
control, and one variant of the router adds this choice to the grid of $k$. The router and the cascade return the
recalibrated vectors of the model that answers. The main variant compares requests on
five numeric features stored in the corpus (prompt length, number of questions, mean and largest number of options,
share of yes/no questions); in \num[n]{ops.suites.main35.features.at_most_one} of the
\num[n]{ops.suites.main35.features.benchmarks} headline benchmarks only one of them varies. The other
representations are computed locally from the request text: sentence embeddings of a small encoder
(\literal{all-MiniLM-L6-v2}); word and bigram TF-IDF reduced by a truncated SVD to at most \literal{128} dimensions;
and the mean of the last hidden states of a small decoder language model (\literal{SmolLM2-360M-Instruct}, first
\literal{256} tokens) reduced by PCA to at most \literal{128} dimensions. All are scaled to unit length, and the
transforms read no label.

\paragraph{Ridge router.}
A ridge regression with an unpenalised intercept predicts each model's mean log score on a request from the
request's sentence embedding; it is fitted on the fitting requests of the benchmark, with the penalty chosen from a
grid of five on the validation fold, and routes like the nearest-neighbour router, with the same cost penalty and
cost cap. The same code reproduces the stored nearest-neighbour router exactly.

\paragraph{Routers that are not told the benchmark.}
These routers pool the fitting requests of all headline benchmarks: one recalibration per model, one setting of the
router and one budget serve all benchmarks, numeric features are standardised on the pooled fitting requests, and
neighbours may come from any benchmark. The single-model reference of this setting is the model with the lowest
pooled log loss on the training folds.

\paragraph{The whole request.}
A second sentence-encoder representation embeds the state, the instructions and the answer options of every
question, with no cut: the text is split into windows of the encoder's length, and the window vectors are averaged by
their numbers of tokens.

\paragraph{Confidence cascade.}
Following \citet{jitkrittum2023when}, the cascade takes the models that no other model beats in both cost and loss
on the fitting folds, keeps at most three of them (the cheapest, the middle one and the one with the lowest loss)
and calls them in that order until the lowest top probability among the request's questions reaches a threshold
$\tau$; a threshold above one never stops early. The cascade pays for every model that it calls. A variant stops
on the mean top probability.

\paragraph{Certified early exit.}
Early exit calls the members of the stacked panel (stacking without a final temperature) in order of decreasing
weight and stops as soon as the uncalled weight cannot change any decoded answer of the request. Its output is the pool of the members called so far.
\section{Additional Results}
\label{sec:full-results}

\subsection{All Methods}
\label{sec:methods-appendix}
Table~\ref{tab:main-results} in the main text gives the methods with their cost and latency.
Table~\ref{tab:benchmark-metrics} adds the gains of LLMRouterBench for every method, and
Tables~\ref{tab:contrasts} and~\ref{tab:contrasts-accuracy} the planned and further contrasts with both intervals,
in log loss and in accuracy.

\begin{table*}[tp]
\centering\footnotesize\setlength{\tabcolsep}{4pt}
\begin{tabular}{@{}lrrrrrr@{}}
\toprule
Method & AvgAcc & Log loss & Gain@$\mathcal{R}$ & Gain@$\mathcal{B}$ & Gain@$\mathcal{B}_d$ & 95\% interval \\
\midrule
Best model per benchmark & 86.26 & 0.384 & $+12.51$ & $+6.24$ & $+0.00$ & -- \\
Random model & 77.81 & 0.671 & $+0.00$ & ${-}4.94$ & ${-}10.19$ & [${-}10.70$, ${-}9.70$] \\
Equal-weight pool & 83.55 & 0.494 & $+8.40$ & $+2.23$ & ${-}3.37$ & [${-}3.79$, ${-}2.95$] \\
Majority vote & 83.20 & 1.708 & $+7.91$ & $+1.80$ & ${-}3.76$ & [${-}4.18$, ${-}3.35$] \\
Skill-weighted pool & 83.92 & 0.475 & $+8.91$ & $+2.69$ & ${-}2.94$ & [${-}3.36$, ${-}2.52$] \\
Equal-weight pool + temperature & 83.55 & 0.426 & $+8.40$ & $+2.23$ & ${-}3.37$ & [${-}3.79$, ${-}2.95$] \\
Stacking & 87.22 & 0.360 & $+13.92$ & $+7.50$ & $+1.23$ & [$+1.01$, $+1.46$] \\
Stacking + temperature & 87.22 & 0.350 & $+13.92$ & $+7.50$ & $+1.23$ & [$+1.01$, $+1.46$] \\
Greedy selection & 86.98 & 0.365 & $+13.66$ & $+7.25$ & $+0.98$ & [$+0.81$, $+1.16$] \\
kNN router, request features & 85.45 & 0.399 & $+11.22$ & $+5.02$ & ${-}1.06$ & [${-}1.37$, ${-}0.76$] \\
kNN router, text embeddings$^{\dagger}$ & 85.69 & 0.393 & $+11.63$ & $+5.43$ & ${-}0.74$ & [${-}1.04$, ${-}0.45$] \\
Confidence cascade & 82.98 & 0.463 & $+7.51$ & $+1.48$ & ${-}4.22$ & [${-}4.60$, ${-}3.86$] \\
Certified early exit & 87.22 & 0.365 & $+13.92$ & $+7.50$ & $+1.23$ & [$+1.01$, $+1.46$] \\
\midrule
Best Single $\mathcal{B}$ (test folds) & 81.90 & -- & -- & -- & ${-}5.29$ & -- \\
\bottomrule
\end{tabular}
\caption{Benchmark metrics of the methods of Table~\ref{tab:main-results}, in percent except log loss: gain over the random model $\mathcal{R}$, Best Single $\mathcal{B}$ and the best model per benchmark $\mathcal{B}_d$, with the 95\% interval of Gain@$\mathcal{B}_d$ (resampling within benchmarks). Gain@$b$ is the mean over benchmarks of the accuracy ratio to $b$, minus one, as in LLMRouterBench.}
\label{tab:benchmark-metrics}
\end{table*}
\begin{table*}[tp]
\centering\scriptsize\setlength{\tabcolsep}{4pt}
\begin{tabular}{@{}llrrrrrr@{}}
\toprule
 & A $-$ B & $\Delta$ log loss & Within benchmarks & Benchmarks resampled & $p$ (Holm) & $\Delta$ acc.\ (pts) & Better / worse \\
\midrule
C1 & Pool $-$ Best ($\mathcal{B}_d$) & $+0.1095$ & [$+0.1040$, $+0.1147$] & [$+0.079$, $+0.143$] & 0.0035 & ${-}2.71$ & 1 / 34 \\
C2 & Pool+T $-$ Pool & ${-}0.0676$ & [${-}0.0713$, ${-}0.0641$] & [${-}0.106$, ${-}0.039$] & 0.0035 & 0 & 33 / 2 \\
C3 & Skill pool $-$ Pool & ${-}0.0188$ & [${-}0.0191$, ${-}0.0185$] & [${-}0.033$, ${-}0.008$] & 0.0035 & $+0.37$ & 30 / 4 \\
C4 & Stack $-$ Pool+T & ${-}0.0668$ & [${-}0.0707$, ${-}0.0627$] & [${-}0.086$, ${-}0.049$] & 0.0035 & $+3.67$ & 35 / 0 \\
C5 & Stack+T $-$ Best ($\mathcal{B}_d$) & ${-}0.0345$ & [${-}0.0371$, ${-}0.0318$] & [${-}0.049$, ${-}0.023$] & 0.0035 & $+0.96$ & 34 / 1 \\
C8 & Router $-$ Best ($\mathcal{B}_d$) & $+0.0144$ & [$+0.0106$, $+0.0180$] & [$+0.007$, $+0.022$] & 0.0035 & ${-}0.81$ & 11 / 24 \\
C9 & Cascade $-$ Best ($\mathcal{B}_d$) & $+0.0785$ & [$+0.0732$, $+0.0842$] & [$+0.051$, $+0.110$] & 0.0035 & ${-}3.28$ & 3 / 32 \\
\midrule
D1 & Greedy $-$ Best ($\mathcal{B}_d$) & ${-}0.0190$ & [${-}0.0212$, ${-}0.0170$] & [${-}0.028$, ${-}0.011$] & -- & $+0.72$ & 31 / 2 \\
D2 & Greedy $-$ Stack+T & $+0.0155$ & [$+0.0134$, $+0.0175$] & [$+0.010$, $+0.023$] & -- & ${-}0.24$ & 1 / 34 \\
D3 & Exit $-$ Stack & $+0.0054$ & [$+0.0039$, $+0.0070$] & [$+0.003$, $+0.008$] & -- & 0 & 5 / 30 \\
D4 & Router (text) $-$ Best ($\mathcal{B}_d$) & $+0.0084$ & [$+0.0040$, $+0.0127$] & [${-}0.001$, $+0.017$] & -- & ${-}0.57$ & 11 / 23 \\
D5 & Router (text) $-$ Router & ${-}0.0060$ & [${-}0.0105$, ${-}0.0016$] & [${-}0.015$, $+0.001$] & -- & $+0.24$ & 17 / 18 \\
\bottomrule
\end{tabular}
\caption{Paired contrasts on the headline suite: A minus B, so a negative log-loss difference favours A. Both intervals are 95\% percentile intervals from 2,000 bootstrap draws: one resamples the source groups within each benchmark, the other resamples the benchmarks. Better / worse counts the benchmarks on which A has the lower / higher log loss (ties are left out). Pool: equal-weight pool; +T: with a temperature; Stack: stacking; Exit: early exit. The contrasts C1 to C5, C8 and C9 were fixed before the first run; their $p$-values (within-benchmark resampling, Holm-adjusted) are descriptive, and 0.0035 is the smallest value that these draws allow. D1 to D5 were added later and are not adjusted. An accuracy difference of 0 is zero by construction.}
\label{tab:contrasts}
\end{table*}
\begin{table*}[tp]
\centering\scriptsize\setlength{\tabcolsep}{3pt}
\begin{tabular}{@{}llrrrr@{}}
\toprule
 & A $-$ B & $\Delta$ acc.\ (pts) & Within benchmarks & Benchmarks resampled & Better / worse \\
\midrule
C1 & Pool $-$ Best ($\mathcal{B}_d$) & ${-}2.71$ & [${-}3.04$, ${-}2.38$] & [${-}4.27$, ${-}1.38$] & 5 / 30 \\
C2 & Pool+T $-$ Pool & 0 & -- & -- & 0 / 0 \\
C3 & Skill pool $-$ Pool & $+0.37$ & [$+0.29$, $+0.45$] & [$+0.21$, $+0.54$] & 25 / 5 \\
C4 & Stack $-$ Pool+T & $+3.67$ & [$+3.36$, $+3.96$] & [$+2.42$, $+5.12$] & 35 / 0 \\
C5 & Stack+T $-$ Best ($\mathcal{B}_d$) & $+0.96$ & [$+0.79$, $+1.12$] & [$+0.53$, $+1.51$] & 32 / 1 \\
C8 & Router $-$ Best ($\mathcal{B}_d$) & ${-}0.81$ & [${-}1.05$, ${-}0.58$] & [${-}1.33$, ${-}0.35$] & 8 / 27 \\
C9 & Cascade $-$ Best ($\mathcal{B}_d$) & ${-}3.28$ & [${-}3.56$, ${-}3.01$] & [${-}5.04$, ${-}1.77$] & 8 / 26 \\
\midrule
D1 & Greedy $-$ Best ($\mathcal{B}_d$) & $+0.72$ & [$+0.59$, $+0.85$] & [$+0.42$, $+1.09$] & 31 / 1 \\
D2 & Greedy $-$ Stack+T & ${-}0.24$ & [${-}0.38$, ${-}0.09$] & [${-}0.53$, ${-}0.04$] & 12 / 20 \\
D3 & Exit $-$ Stack & 0 & -- & -- & 0 / 0 \\
D4 & Router (text) $-$ Best ($\mathcal{B}_d$) & ${-}0.57$ & [${-}0.80$, ${-}0.35$] & [${-}0.97$, ${-}0.19$] & 7 / 25 \\
D5 & Router (text) $-$ Router & $+0.24$ & [${-}0.02$, $+0.52$] & [${-}0.08$, $+0.58$] & 19 / 16 \\
\bottomrule
\end{tabular}
\caption{The paired contrasts in accuracy, in percentage points, with the same two kinds of 95\% interval. Better / worse counts the benchmarks on which A is more / less accurate. A difference of 0 is zero by construction.}
\label{tab:contrasts-accuracy}
\end{table*}

\paragraph{Other suites.}
Tables~\ref{tab:sensitivity-loss} and~\ref{tab:sensitivity-accuracy} repeat the comparison on
\num[n]{facts.sensitivity.other_suites} parts of the headline suite: without the two recall-only benchmarks, without
the largest one, and without the four on which the mean model accuracy exceeds \literal{95}\%. The order of the
methods by accuracy is unchanged, and the order by log loss differs by one swap of neighbours at most. Stacking's
lead over $\mathcal{B}_d$ does not depend on the recall-only benchmarks
(\num[f2]{facts.recall_only.stack_final_temperature.headline} points with them and
\num[f2]{facts.recall_only.stack_final_temperature.without} without).

\begin{table*}[tp]
\centering\footnotesize
\begin{tabular}{@{}lrrrr@{}}
\toprule
Method & Headline (35) & No recall-only (33) & No largest (34) & No saturated (31) \\
\midrule
Best model per benchmark & 0.384 & 0.384 & 0.395 & 0.431 \\
Random model & 0.671 & 0.656 & 0.689 & 0.748 \\
Equal-weight pool & 0.494 & 0.485 & 0.507 & 0.549 \\
Majority vote & 1.708 & 1.567 & 1.757 & 1.922 \\
Skill-weighted pool & 0.475 & 0.469 & 0.487 & 0.528 \\
Equal-weight pool + temperature & 0.426 & 0.423 & 0.438 & 0.477 \\
Stacking & 0.360 & 0.359 & 0.370 & 0.404 \\
Stacking + temperature & 0.350 & 0.349 & 0.360 & 0.393 \\
Greedy selection & 0.365 & 0.365 & 0.376 & 0.410 \\
kNN router, request features & 0.399 & 0.400 & 0.410 & 0.447 \\
kNN router, text embeddings$^{\dagger}$ & 0.393 & 0.396 & 0.404 & 0.440 \\
Confidence cascade & 0.463 & 0.462 & 0.475 & 0.519 \\
Certified early exit & 0.365 & 0.365 & 0.375 & 0.410 \\
\bottomrule
\end{tabular}
\caption{Log loss on every suite (number of benchmarks in parentheses).}
\label{tab:sensitivity-loss}
\end{table*}
\begin{table*}[tp]
\centering\footnotesize
\begin{tabular}{@{}lrrrr@{}}
\toprule
Method & Headline (35) & No recall-only (33) & No largest (34) & No saturated (31) \\
\midrule
Best model per benchmark & 86.3 & 86.5 & 85.9 & 84.6 \\
Random model & 77.8 & 78.6 & 77.2 & 75.3 \\
Equal-weight pool & 83.6 & 84.1 & 83.1 & 81.6 \\
Majority vote & 83.2 & 83.7 & 82.7 & 81.2 \\
Skill-weighted pool & 83.9 & 84.4 & 83.5 & 82.0 \\
Equal-weight pool + temperature & 83.6 & 84.1 & 83.1 & 81.6 \\
Stacking & 87.2 & 87.5 & 86.9 & 85.6 \\
Stacking + temperature & 87.2 & 87.5 & 86.9 & 85.6 \\
Greedy selection & 87.0 & 87.2 & 86.6 & 85.4 \\
kNN router, request features & 85.5 & 85.8 & 85.0 & 83.7 \\
kNN router, text embeddings$^{\dagger}$ & 85.7 & 86.0 & 85.3 & 83.9 \\
Confidence cascade & 83.0 & 83.5 & 82.5 & 80.9 \\
Certified early exit & 87.2 & 87.5 & 86.9 & 85.6 \\
\bottomrule
\end{tabular}
\caption{Accuracy (\%) on every suite.}
\label{tab:sensitivity-accuracy}
\end{table*}

\subsection{Models}
\label{sec:models-appendix}
Table~\ref{tab:models} gives each model on the headline suite, and Tables~\ref{tab:model-accuracy-1}
and~\ref{tab:model-accuracy-2} every model on every benchmark.

\begin{table*}[tp]
\centering\footnotesize
\begin{tabular}{@{}lrrrrr@{}}
\toprule
Model & Accuracy & Log loss & Calibration & USD per & Median \\
 & (\%) & & error & 1,000 requests & latency (s) \\
\midrule
\texttt{liquid/d1-ultra} & 81.9 & 0.499 & 0.050 & 0.073 & 0.41 \\
\texttt{liquid/d1.2-ultra} & 81.4 & 0.497 & 0.052 & 0.073 & 0.48 \\
\texttt{perplexity/pplx-decider-v1.2} & 81.4 & 0.542 & 0.055 & 0.036 & 0.44 \\
\texttt{nace-ai/drex-max-finance} & 81.3 & 0.528 & 0.065 & 0.109 & 0.95 \\
\texttt{liquid/d1.2} & 80.3 & 0.532 & 0.056 & 0.035 & 0.41 \\
\texttt{liquid/d1.1} & 79.3 & 0.552 & 0.049 & 0.035 & 0.36 \\
\texttt{cloudflare/clef} & 77.5 & 0.649 & 0.062 & 0.405 & 0.90 \\
\texttt{nace-ai/drex-max} & 77.1 & 0.653 & 0.095 & 0.131 & 0.55 \\
\texttt{microsoft/microsoft-decision-1.1} & 76.9 & 0.687 & 0.078 & 0.079 & 0.39 \\
\texttt{typesafe/jev-1.14} & 76.8 & 0.925 & 0.066 & 0.097 & 0.33 \\
\texttt{liquid/d1} & 76.2 & 0.664 & 0.058 & 0.035 & 0.38 \\
\texttt{cloudflare/clef-omni} & 75.8 & 0.732 & 0.082 & 0.315 & 0.32 \\
\texttt{typesafe/jev-1.13} & 75.2 & 1.067 & 0.079 & 0.097 & 0.33 \\
\texttt{perplexity/pplx-decider-flash-v1.2} & 75.2 & 0.704 & 0.080 & 0.018 & 0.33 \\
\texttt{nace-ai/drex-v2.0} & 74.9 & 0.649 & 0.085 & 0.105 & 0.57 \\
\texttt{cloudflare/clef-flash} & 74.6 & 0.772 & 0.084 & 0.042 & 0.70 \\
\texttt{nace-ai/drex-v1.5} & 72.4 & 0.836 & 0.118 & 0.066 & 0.49 \\
\bottomrule
\end{tabular}
\caption{The models on the headline suite (equal weight per benchmark), by accuracy. Each model is scored as a panel of one: normalised vectors, first maximum. Calibration error is the expected calibration error with ten bins. Cost and median latency are means over the benchmarks of the suite.}
\label{tab:models}
\end{table*}
\begin{table*}[tp]
\centering\tiny\setlength{\tabcolsep}{2.4pt}
\begin{tabular}{@{}lrrrrrrrrrrrrrrrrrr@{}}
\toprule
 & \multicolumn{10}{c}{Knowledge} & \multicolumn{8}{c}{Language} \\
\cmidrule(lr){2-11}\cmidrule(lr){12-19}
Model & GPQA & Wino & GSM8K & Chess & MuSR & SATA & CRUX & CLad & MMLUP & BBH & CNLI & ANLI & HSwag & ACOS & FinE & iSarc & VAST & NLI4CT \\
\midrule
\textbf{Clef} & 46.9 & 93.4 & 59.9 & 24.4 & \cellcolor{secondcell}84.0 & 84.5 & 86.0 & 94.0 & 66.0 & 73.9 & 86.7 & 69.3 & 98.1 & 97.7 & 96.2 & 95.6 & 59.2 & 84.5 \\
\textbf{Clef-Flash} & 50.5 & \cellcolor{bestcell}97.5 & 46.4 & 23.0 & \cellcolor{bestcell}86.6 & 84.7 & \cellcolor{secondcell}87.0 & \cellcolor{bestcell}97.7 & 65.2 & 68.8 & 87.9 & 59.2 & \cellcolor{secondcell}98.6 & 96.0 & \cellcolor{bestcell}97.0 & \cellcolor{bestcell}96.4 & 49.6 & 79.8 \\
\textbf{Clef-Omni} & 46.4 & \cellcolor{secondcell}96.8 & 68.2 & 21.8 & 83.9 & 83.0 & 85.1 & \cellcolor{secondcell}97.4 & 69.3 & 68.6 & 87.3 & 63.4 & \cellcolor{bestcell}98.9 & 95.2 & \cellcolor{secondcell}96.7 & \cellcolor{secondcell}96.1 & 50.9 & 79.8 \\
\textbf{D1} & 51.0 & 90.0 & 45.8 & 19.5 & 60.9 & 85.0 & 68.8 & 70.2 & 66.6 & 90.2 & \cellcolor{secondcell}89.0 & 74.8 & 97.2 & \cellcolor{secondcell}99.4 & 90.3 & 94.9 & 79.6 & 84.1 \\
\textbf{D1-Ultra} & 52.6 & 93.4 & \cellcolor{bestcell}83.7 & 22.7 & 71.0 & 85.9 & \cellcolor{bestcell}88.1 & 86.4 & 74.6 & 90.3 & 84.9 & \cellcolor{bestcell}79.7 & 96.7 & 99.1 & 94.9 & 93.0 & 79.2 & \cellcolor{secondcell}88.6 \\
\textbf{D1.1} & 54.1 & 90.4 & 73.7 & 18.3 & 61.7 & 85.1 & 77.5 & 83.0 & 70.9 & 88.5 & 85.1 & 76.8 & 96.1 & 98.1 & 92.8 & 92.6 & 79.4 & 87.6 \\
\textbf{D1.2} & 58.2 & 91.5 & 75.5 & 18.9 & 67.7 & 86.3 & 78.4 & 85.4 & 70.9 & 91.6 & 87.2 & 77.7 & 96.4 & 98.4 & 92.7 & 92.9 & 79.2 & 87.8 \\
\textbf{D1.2-Ultra} & 55.6 & 92.7 & 80.8 & 19.8 & 68.2 & \cellcolor{secondcell}87.4 & 84.0 & 84.9 & 72.5 & 93.1 & 87.5 & \cellcolor{secondcell}79.7 & 96.9 & 97.6 & 94.6 & 94.0 & \cellcolor{bestcell}81.8 & \cellcolor{bestcell}89.1 \\
\textbf{MS-1.1} & 51.5 & 87.8 & 81.2 & 22.3 & 69.4 & 82.8 & 79.8 & 84.3 & 70.2 & 78.8 & 81.9 & 78.8 & 93.5 & 94.6 & 89.7 & 85.6 & 71.7 & 86.1 \\
\textbf{Drex-Max} & 56.6 & 94.6 & 83.3 & \cellcolor{secondcell}24.5 & 69.7 & 85.3 & 85.4 & 85.6 & 72.5 & \cellcolor{secondcell}93.4 & 86.5 & 77.2 & 97.9 & 98.2 & 95.1 & 92.7 & 75.9 & 87.3 \\
\textbf{Drex-Max-Fin} & 54.1 & 93.7 & 82.1 & \cellcolor{bestcell}25.3 & 75.8 & \cellcolor{bestcell}88.4 & 84.0 & 92.3 & 76.6 & \cellcolor{bestcell}93.9 & \cellcolor{bestcell}89.4 & 76.4 & 96.8 & 98.7 & 84.0 & 92.9 & \cellcolor{secondcell}81.5 & 87.6 \\
\textbf{Drex-1.5} & 43.9 & 87.6 & 54.9 & 22.0 & 57.7 & 80.4 & 76.1 & 86.1 & 58.7 & 85.6 & 87.6 & 70.6 & 92.1 & \cellcolor{bestcell}99.5 & 88.3 & 91.5 & 77.4 & 81.6 \\
\textbf{Drex-2.0} & 51.5 & 86.1 & 66.8 & 23.4 & 59.6 & 80.3 & 84.6 & 84.5 & 72.8 & 78.4 & 87.4 & 70.5 & 95.9 & 98.0 & 91.0 & 93.2 & 78.4 & 83.2 \\
\textbf{Decider-Flash} & 43.9 & 88.6 & 67.5 & 21.3 & 59.2 & 73.6 & 62.8 & 85.0 & 59.4 & 73.1 & 87.9 & 68.4 & 93.8 & 97.4 & 91.2 & 93.5 & 77.4 & 83.7 \\
\textbf{Decider} & 54.1 & 94.5 & \cellcolor{secondcell}83.4 & 24.4 & 70.5 & 84.8 & 84.9 & 90.8 & 72.1 & 85.1 & 88.3 & 75.1 & 96.6 & 98.6 & 95.0 & 93.1 & 80.6 & 86.9 \\
\textbf{Jev-1.13} & \cellcolor{bestcell}77.6 & 91.6 & 72.1 & 17.2 & 65.4 & 81.1 & 72.8 & 72.7 & \cellcolor{bestcell}82.8 & 93.0 & 76.8 & 74.6 & 94.5 & 96.9 & 86.6 & 75.7 & 64.6 & 86.2 \\
\textbf{Jev-1.14} & \cellcolor{secondcell}73.5 & 91.0 & 73.2 & 17.1 & 71.9 & 82.0 & 74.6 & 88.5 & \cellcolor{secondcell}82.3 & 93.1 & 75.5 & 74.0 & 94.3 & 97.1 & 87.8 & 77.1 & 73.0 & 87.0 \\
\midrule
Best ($\mathcal{B}_d$) & 77.6 & 97.1 & 83.1 & 24.0 & 86.6 & 88.4 & 86.0 & 97.7 & 82.8 & 93.7 & 88.5 & 79.7 & 98.9 & 99.4 & 96.8 & 96.1 & 81.8 & 88.7 \\
Stack+T & 77.6 & 97.7 & 87.6 & 25.2 & 88.2 & 88.7 & 93.3 & 98.0 & 83.9 & 94.4 & 89.9 & 81.8 & 99.1 & 99.5 & 97.7 & 96.6 & 82.2 & 89.5 \\
Router & 72.4 & 96.8 & 83.0 & 23.8 & 87.1 & 88.9 & 86.8 & 97.5 & 80.9 & 95.5 & 89.1 & 79.1 & 98.3 & 99.4 & 96.1 & 96.2 & 80.3 & 87.1 \\
Cascade & 71.9 & 97.2 & 77.0 & 21.6 & 67.6 & 76.0 & 78.2 & 93.0 & 83.1 & 93.5 & 87.9 & 76.6 & 99.2 & 98.1 & 96.3 & 96.4 & 81.4 & 89.1 \\
\bottomrule
\end{tabular}
\caption{Accuracy (\%) of every model on every benchmark, and of four methods below the rule. Shaded: the best and second-best model of each benchmark.}
\label{tab:model-accuracy-1}
\end{table*}

\begin{table*}[tp]
\centering\tiny\setlength{\tabcolsep}{2.4pt}
\begin{tabular}{@{}lrrrrrrrrrrrrrrrrr@{}}
\toprule
 & \multicolumn{6}{c}{Retrieval} & \multicolumn{5}{c}{Tools} & \multicolumn{6}{c}{Arts} \\
\cmidrule(lr){2-7}\cmidrule(lr){8-12}\cmidrule(lr){13-18}
Model & BK77 & CLINC & BRIGHT & ESCI & Phish & HoVer & BFCL & TRet & APIB & Home & W2C & BPoMP & Humi & POP & cfcol & Haber & NYC \\
\midrule
\textbf{Clef} & \cellcolor{secondcell}94.0 & \cellcolor{bestcell}97.3 & \cellcolor{bestcell}69.2 & 68.2 & 79.0 & 65.6 & 99.3 & 94.3 & \cellcolor{secondcell}91.5 & 98.9 & 72.6 & 96.9 & 66.3 & 15.2 & 65.7 & 68.4 & 68.9 \\
\textbf{Clef-Flash} & 90.7 & 60.9 & 60.0 & 68.0 & 75.4 & 60.9 & \cellcolor{bestcell}99.5 & 89.5 & \cellcolor{bestcell}93.3 & \cellcolor{bestcell}99.9 & 64.8 & 95.4 & \cellcolor{bestcell}75.5 & 1.5 & 65.7 & \cellcolor{bestcell}71.4 & 67.2 \\
\textbf{Clef-Omni} & \cellcolor{bestcell}94.2 & \cellcolor{secondcell}96.8 & \cellcolor{secondcell}67.8 & 68.6 & 72.5 & 62.8 & \cellcolor{secondcell}99.4 & 90.4 & 86.8 & 97.8 & 57.0 & \cellcolor{secondcell}97.0 & \cellcolor{secondcell}71.1 & 2.6 & 63.6 & \cellcolor{secondcell}70.0 & 65.5 \\
\textbf{D1} & 88.9 & 91.5 & 57.5 & 73.3 & 71.9 & 75.1 & 98.6 & 95.2 & 83.9 & 95.9 & 86.9 & 92.6 & 65.4 & 30.9 & \cellcolor{bestcell}71.2 & 59.1 & 72.0 \\
\textbf{D1-Ultra} & 87.2 & 92.2 & 63.7 & 73.6 & 80.8 & 87.1 & 98.2 & 96.2 & 88.6 & 97.4 & 89.7 & 95.1 & 63.7 & \cellcolor{secondcell}79.3 & 68.1 & 59.3 & \cellcolor{bestcell}81.6 \\
\textbf{D1.1} & 88.0 & 90.4 & 49.5 & 72.2 & 74.4 & 85.6 & 98.6 & 95.1 & 86.8 & 97.2 & 88.3 & 93.0 & 62.3 & 74.4 & 68.0 & 61.9 & 77.1 \\
\textbf{D1.2} & 89.2 & 92.9 & 43.2 & 72.8 & 76.1 & 87.7 & 98.4 & 95.4 & 86.8 & 96.3 & 91.1 & 93.9 & 64.2 & 76.4 & 69.1 & 62.2 & 76.7 \\
\textbf{D1.2-Ultra} & 91.8 & 93.8 & 43.2 & \cellcolor{secondcell}73.9 & 74.5 & \cellcolor{bestcell}88.6 & 99.0 & 92.1 & 87.4 & 98.4 & \cellcolor{bestcell}92.2 & 95.5 & 65.6 & \cellcolor{bestcell}79.8 & \cellcolor{secondcell}71.0 & 64.3 & 78.6 \\
\textbf{MS-1.1} & 83.0 & 92.3 & 56.6 & 71.0 & 82.6 & 77.9 & 98.4 & 91.9 & 86.4 & 95.5 & 88.4 & 92.1 & 64.1 & 35.9 & 64.7 & 44.0 & 77.8 \\
\textbf{Drex-Max} & 90.2 & 94.3 & 23.3 & 73.8 & 74.8 & 87.7 & 99.2 & 7.2 & 88.4 & 97.5 & 85.2 & 95.2 & 63.1 & 59.7 & 66.6 & 50.5 & 80.1 \\
\textbf{Drex-Max-Fin} & 93.0 & 93.8 & 48.2 & \cellcolor{bestcell}74.2 & \cellcolor{bestcell}91.3 & \cellcolor{secondcell}88.2 & 98.8 & 52.0 & 87.8 & \cellcolor{secondcell}99.7 & \cellcolor{secondcell}91.7 & \cellcolor{bestcell}98.1 & 67.0 & \cellcolor{bestcell}79.8 & 65.5 & 61.6 & \cellcolor{secondcell}80.5 \\
\textbf{Drex-1.5} & 85.7 & 90.5 & 11.6 & 72.2 & 66.5 & 86.2 & 98.0 & 1.0 & 85.6 & 93.5 & 89.5 & 82.1 & 63.0 & 76.5 & 58.4 & 59.5 & 73.5 \\
\textbf{Drex-2.0} & 81.1 & 88.4 & 23.1 & 72.3 & 75.0 & 85.8 & 98.7 & 14.8 & 83.3 & 96.0 & 86.5 & 88.9 & 63.5 & \cellcolor{bestcell}79.8 & 59.2 & 62.5 & 77.7 \\
\textbf{Decider-Flash} & 90.4 & 91.5 & 50.5 & 72.1 & 72.7 & 86.3 & 98.2 & 90.2 & 86.0 & 98.1 & 89.4 & 90.4 & 61.3 & 24.9 & 66.3 & 61.3 & 74.2 \\
\textbf{Decider} & 91.0 & 92.7 & 67.4 & 73.3 & 77.0 & 86.4 & 98.8 & \cellcolor{bestcell}96.8 & 87.6 & 98.9 & 90.1 & 95.0 & 65.2 & 61.5 & 68.0 & 59.2 & 79.7 \\
\textbf{Jev-1.13} & 80.6 & 88.8 & 66.9 & 66.5 & 63.0 & 73.3 & 98.4 & \cellcolor{secondcell}96.3 & 87.8 & 96.0 & 80.8 & 90.4 & 62.2 & 18.9 & 64.9 & 46.2 & 70.1 \\
\textbf{Jev-1.14} & 80.7 & 89.9 & 65.5 & 68.2 & \cellcolor{secondcell}83.9 & 77.0 & 98.6 & 95.8 & 86.6 & 95.3 & 81.9 & 90.0 & 61.6 & 16.9 & 66.0 & 44.5 & 71.4 \\
\midrule
Best ($\mathcal{B}_d$) & 94.0 & 97.3 & 69.9 & 74.2 & 91.3 & 88.6 & 99.4 & 95.9 & 93.3 & 99.7 & 92.2 & 98.1 & 75.5 & 79.8 & 71.2 & 71.4 & 80.5 \\
Stack+T & 94.4 & 97.6 & 69.2 & 74.8 & 91.6 & 89.1 & 99.5 & 96.3 & 94.1 & 99.7 & 92.4 & 98.5 & 75.7 & 83.2 & 71.7 & 72.0 & 82.0 \\
Router & 93.2 & 96.6 & 63.9 & 73.2 & 91.3 & 88.0 & 99.3 & 95.4 & 92.5 & 99.9 & 91.2 & 96.9 & 72.5 & 80.0 & 68.4 & 70.3 & 79.7 \\
Cascade & 94.0 & 97.1 & 57.1 & 74.1 & 84.9 & 88.8 & 99.3 & 92.6 & 90.6 & 99.7 & 92.1 & 98.1 & 61.3 & 80.3 & 68.0 & 62.7 & 79.5 \\
\bottomrule
\end{tabular}
\caption{Accuracy (\%) of every model on every benchmark, and of four methods below the rule. Shaded: the best and second-best model of each benchmark. (continued)}
\label{tab:model-accuracy-2}
\end{table*}

\begin{figure}[t]
\centering
\includegraphics{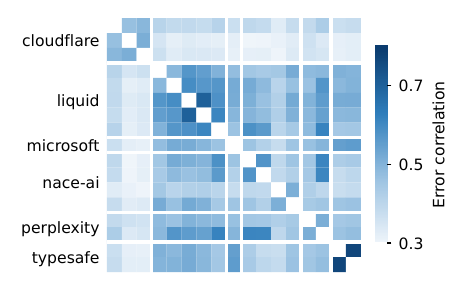}
\caption{\assetcaption{similarity}}
\label{fig:similarity}
\end{figure}

\paragraph{Error correlation given difficulty.}
The errors of two models correlate whenever some questions are hard for every model. To set this apart, we group the
judged questions of a benchmark by the number of the \emph{other} models that answer them wrongly, and compute the
correlation of the two models' errors within these groups. The mean over pairs and benchmarks falls from
\num[f2]{homog.suites.main35.all_pairs.error_correlation} to
\num[f2]{homog.suites.main35.all_pairs.error_correlation_given_others}. It is
\num[f2]{homog.suites.main35.same_organisation.error_correlation_given_others} for pairs from one provider and
\num[f2]{homog.suites.main35.different_organisation.error_correlation_given_others} for pairs from different
providers (Figure~\ref{fig:similarity} shows the unadjusted correlations). The number of other models that err
is an imperfect proxy for difficulty, so we interpret only the difference between the two groups of pairs.

\paragraph{Other measures of similarity.}
On the \num[n]{homog.suites.main35.all_pairs.both_wrong_agreement_benchmarks} benchmarks with at least three
options, two models that are both wrong choose the same wrong option in
\num[pct0]{homog.suites.main35.all_pairs.both_wrong_agreement} of cases on average over benchmarks, against
\num[pct0]{homog.suites.main35.all_pairs.chance_agreement} under a uniform choice among the wrong options, the
measure of \citet{kim2025correlated}. Two models fail the same question
\num[f1]{homog.suites.main35.all_pairs.joint_failure_ratio} times as often as independent failures would
imply~\citep{bommasani2022picking}. Under an equal-correlation approximation the models of a benchmark are worth
\num[f1]{diag.suites.main35.overall.headroom.0.effective_voters} independent voters on average (between
\num[f1]{facts.homogeneity.effective_voters_min} and \num[f1]{facts.homogeneity.effective_voters_max}). On the
\num[n]{homog.suites.main35.all_models.none_right_with_choice} judged questions with at least two wrong options that
no model answers correctly, all models choose the same wrong option in
\num[pct0]{homog.suites.main35.all_models.none_right_unanimous} of cases on average over benchmarks. Such questions
are candidates for an error in the gold answer or for one dominant distractor; we did not inspect them.

\subsection{Panel Size, Calibration and Label Budgets}
\label{sec:panels-appendix}

\begin{table*}[tp]
\centering\scriptsize\setlength{\tabcolsep}{4pt}
\begin{tabular}{@{}llrrrrrrr@{}}
\toprule
Policy & Rule & Log loss & Acc.\ (\%) & Own (\%) & Cost & Calls & $\Delta$ log loss to best per benchmark & Cap binds \\
\midrule
Per-benchmark control & capped & 0.3848 & 86.14 & 78.3 & 85\% & 1.00 & $+0.0003$ [${-}0.0018$, $+0.0023$] & 6 / 175 \\
 & uncapped & 0.3842 & 86.19 & 78.3 & 97\% & 1.00 & ${-}0.0003$ [${-}0.0023$, $+0.0018$] &  \\
kNN router, request features & capped & 0.3988 & 85.45 & 79.0 & 63\% & 1.00 & $+0.0144$ [$+0.0106$, $+0.0180$] & 71 / 175 \\
 & uncapped & 0.3944 & 85.80 & 79.6 & 86\% & 1.00 & $+0.0099$ [$+0.0061$, $+0.0139$] &  \\
kNN router, text embeddings$^{\dagger}$ & capped & 0.3928 & 85.69 & 79.5 & 55\% & 1.00 & $+0.0084$ [$+0.0040$, $+0.0127$] & 69 / 175 \\
 & uncapped & 0.3877 & 86.07 & 80.3 & 69\% & 1.00 & $+0.0033$ [${-}0.0013$, $+0.0077$] &  \\
Cascade, lowest confidence & capped & 0.4630 & 82.98 & 76.6 & 47\% & 1.68 & $+0.0785$ [$+0.0732$, $+0.0842$] & 144 / 175 \\
 & uncapped & 0.3867 & 86.17 & 79.1 & 123\% & 2.62 & $+0.0022$ [${-}0.0006$, $+0.0051$] &  \\
Cascade, mean confidence & capped & 0.4574 & 83.35 & 76.7 & 55\% & 1.70 & $+0.0730$ [$+0.0677$, $+0.0787$] & 146 / 175 \\
 & uncapped & 0.3869 & 86.12 & 78.4 & 113\% & 2.65 & $+0.0025$ [${-}0.0003$, $+0.0052$] &  \\
\bottomrule
\end{tabular}
\caption{Operating points of the router, its per-benchmark control and the cascade on the headline suite. ``Capped'' is the rule of the main table: per benchmark and test fold, the setting with the lowest validation loss among those whose validation cost does not exceed that of the best model per benchmark. ``Uncapped'' takes the lowest validation loss at any cost. Cost is relative to the best model per benchmark (0.3845 nats, 86.26\% accuracy, 78.7\% own metric). Intervals: paired bootstrap over source groups, not adjusted. ``Cap binds'' counts the combinations of benchmark and test fold in which the cap changed the choice. $^{\dagger}$ needs the benchmark text.}
\label{tab:operating-points}
\end{table*}
\begin{table*}[tp]
\centering\footnotesize
\begin{tabular}{@{}lrrrr@{}}
\toprule
 & \multicolumn{2}{c}{Log loss} & \multicolumn{2}{c}{Accuracy (\%)} \\
\cmidrule(lr){2-3}\cmidrule(l){4-5}
Method & Frozen folds & Redrawn folds & Frozen folds & Redrawn folds \\
\midrule
Best model per benchmark & 0.3845 & 0.3843 to 0.3869 & 86.26 & 86.13 to 86.27 \\
Random model & 0.6714 & 0.6667 to 0.6821 & 77.81 & 77.36 to 77.98 \\
Equal-weight pool & 0.4939 & 0.4939 to 0.4939 & 83.55 & 83.55 to 83.55 \\
Majority vote & 1.7084 & 1.7084 to 1.7084 & 83.20 & 83.20 to 83.20 \\
Skill-weighted pool & 0.4751 & 0.4751 to 0.4751 & 83.92 & 83.88 to 83.92 \\
Equal-weight pool + temperature & 0.4263 & 0.4262 to 0.4269 & 83.55 & 83.55 to 83.55 \\
Stacking & 0.3596 & 0.3594 to 0.3608 & 87.22 & 87.14 to 87.33 \\
Stacking + temperature & 0.3500 & 0.3487 to 0.3515 & 87.22 & 87.14 to 87.33 \\
Greedy selection & 0.3654 & 0.3638 to 0.3665 & 86.98 & 86.90 to 87.14 \\
kNN router, request features & 0.3988 & 0.3967 to 0.4028 & 85.45 & 85.28 to 85.75 \\
Confidence cascade & 0.4630 & 0.4527 to 0.4781 & 82.98 & 82.53 to 83.39 \\
Certified early exit & 0.3650 & 0.3645 to 0.3673 & 87.22 & 87.14 to 87.33 \\
\bottomrule
\end{tabular}
\caption{The methods that the simulator fits, on the frozen folds and on 20 redrawn assignments of the source groups to five folds (lowest and highest value over the draws).}
\label{tab:fold-redraws}
\end{table*}

\begin{figure}[t]
\centering
\includegraphics{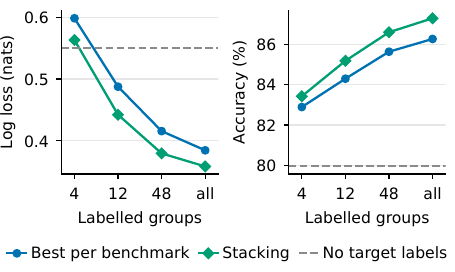}
\caption{\assetcaption{label_budget}}
\label{fig:label-budget}
\end{figure}

\paragraph{Panel size and calibration.}
For the full equal-weight pool (Figure~\ref{fig:panel-size}) a final temperature matters more than calibrating the
members: its log loss is \num[f3]{facts.calibration_order.none} without either,
\num[f3]{facts.calibration_order.calibrated_members} with calibrated members,
\num[f3]{facts.calibration_order.final_temperature} with a final temperature and
\num[f3]{facts.calibration_order.calibrated_members_and_final_temperature} with both.

\paragraph{Label budgets.}
Figure~\ref{fig:label-budget} fits $\mathcal{B}_d$ and the stacked panel on a sample of labelled source groups of
the target benchmark, and compares them with a model chosen on the other task areas without any target label (log
loss \num[f3]{facts.no_target_labels.transfer_single}, accuracy
\num[pct1]{facts.no_target_labels.transfer_single_accuracy}). With four groups both fitted policies have a higher
log loss than that model, and with \num[n]{facts.label_budget.groups_where_selection_beats_transfer} groups both are
ahead of it on both metrics. At every budget the stacked panel has a lower log loss than $\mathcal{B}_d$.

\subsection{Operating Points of the Router and the Cascade}
\label{sec:operating-points}

The router and the cascade each have a setting that trades loss against cost. The main results use the capped rule
of Appendix~\ref{sec:setup}; Table~\ref{tab:operating-points} adds the same policies tuned for validation loss
alone.

\paragraph{Under the cap.}
For the router with numeric features the cap changed the chosen cost weight in
\num[n]{ops.suites.main35.selection.knn_router.cap_changed_choice} of the
\num[n]{ops.suites.main35.selection.knn_router.cells} combinations of benchmark and test fold (cells), and in
\num[n]{ops.suites.main35.selection.knn_router.none_within_budget} cells no value was within the cap. For the
cascade it changed the threshold in \num[n]{ops.suites.main35.selection.confidence_cascade.cap_changed_choice}
cells. A cascade pays for every model that it calls, so a cap at the cost of one model forces it to stop early on
many requests: \num[pct0]{facts.operating_points.cascade_gap_closed_by_lifting_the_cap} of its capped gap in log loss
to $\mathcal{B}_d$ disappears without the cap.

\paragraph{Without the cap.}
Tuned for validation loss alone, the cascade calls
\num[f2]{ops.suites.main35.points.confidence_cascade.uncapped.calls_per_request} models per request and costs more
than $\mathcal{B}_d$. Its log loss can then not be told from that of $\mathcal{B}_d$
(\num[s4]{ops.suites.main35.contrasts.cascade_uncapped_vs_single.log_loss.difference} nats, interval
\num[s4]{ops.suites.main35.contrasts.cascade_uncapped_vs_single.log_loss.ci_low} to
\num[s4]{ops.suites.main35.contrasts.cascade_uncapped_vs_single.log_loss.ci_high}). The nearest-neighbour router stays
cheaper than $\mathcal{B}_d$ and worse (\num[s4]{ops.suites.main35.contrasts.router_uncapped_vs_single.log_loss.difference}
nats, \num[s4]{ops.suites.main35.contrasts.router_uncapped_vs_single.log_loss.ci_low} to
\num[s4]{ops.suites.main35.contrasts.router_uncapped_vs_single.log_loss.ci_high}); for the text router the interval
includes zero (\num[s4]{ops.suites.main35.contrasts.text_router_uncapped_vs_single.log_loss.ci_low} to
\num[s4]{ops.suites.main35.contrasts.text_router_uncapped_vs_single.log_loss.ci_high}). Against its control, the
uncapped nearest-neighbour router is
\num[a3]{ops.suites.main35.contrasts.router_uncapped_vs_control.log_loss.difference} nats worse (interval
\num[s3]{ops.suites.main35.contrasts.router_uncapped_vs_control.log_loss.ci_low} to
\num[s3]{ops.suites.main35.contrasts.router_uncapped_vs_control.log_loss.ci_high}) at
\num[pct0]{facts.operating_points.router_cost_relative_to_control.uncapped} of the control's cost. Where two or more
of the five numeric features vary, the capped router is level with $\mathcal{B}_d$
(\num[s4]{ops.suites.main35.router_by_features.two_or_more.capped_log_loss_difference} nats), against
\num[s4]{ops.suites.main35.router_by_features.at_most_one.capped_log_loss_difference} nats where at most one
varies. No fixed setting of a nearest-neighbour router or of the cascade has a lower log loss than $\mathcal{B}_d$.
A cascade that stops
on the mean top probability behaves alike: it is
\num[pts2]{ops.suites.main35.contrasts.mean_cascade_capped_vs_single.accuracy.difference} points below
$\mathcal{B}_d$ under the cap and
\num[pts2]{ops.suites.main35.contrasts.mean_cascade_uncapped_vs_single.accuracy.difference} points below without it.

\begin{figure*}[t]
\centering
\includegraphics{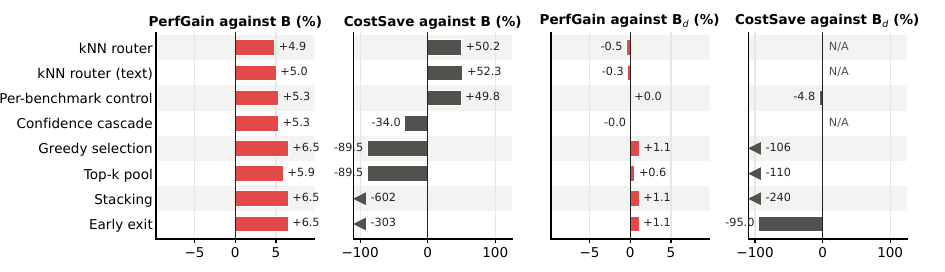}
\caption{\assetcaption{gains}}
\label{fig:gains}
\end{figure*}

\paragraph{Best settings in hindsight.}
Figure~\ref{fig:gains} gives two metrics of LLMRouterBench for a policy with several settings, against Best Single
$\mathcal{B}$ and against $\mathcal{B}_d$: PerfGain, the accuracy of its most accurate setting over the
reference's, and CostSave, the cost saved by its cheapest setting that is at least as accurate as the reference.
Both choose the setting on the test folds, so they show the best case and not a setting that could have been chosen
in advance; every other comparison of this paper tunes on the validation fold.

\paragraph{The benchmarks' own metrics.}
The kit's scorers accept a weight per source group, so the own metric can be resampled like the other metrics. The
capped router is \num[pts2]{fu.own.router_minus_best} points above $\mathcal{B}_d$ on the mean of the own metrics
(interval \num[spts2]{fu.own.router_groups_low} to \num[spts2]{fu.own.router_groups_high} when source groups are
resampled, \num[spts2]{fu.own.router_over_low} to \num[spts2]{fu.own.router_over_high} when benchmarks are). It is
above $\mathcal{B}_d$ on \num[n]{fu.own.router_above} benchmarks and below it on \num[n]{fu.own.router_below}; without
the two benchmarks with the largest gains, iSarcasmEval and ACOS, both scored by an F\literal{1} measure, its mean is
\num[a2]{facts.own_metric.knn_router.points_without_two_largest_gains} points below. $\mathcal{B}_d$ is chosen by log
loss, which on these two benchmarks picks a model with a low F\literal{1}. A reference that picks, for each test fold, the
model with the best own metric on the four other folds reaches \num[pct1]{fu.own.own_reference_own} on the own
metrics, against \num[pct1]{fu.own.reference_own} for $\mathcal{B}_d$, and both routers are below it (by
\num[pts2]{fu.own.router_minus_own} and \num[pts2]{fu.own.text_router_minus_own} points; both intervals exclude zero).
Stacking with a final temperature is \num[pts2]{fu.own.stack_minus_best} points above $\mathcal{B}_d$ on the own
metrics (\num[spts2]{fu.own.stack_groups_low} to \num[spts2]{fu.own.stack_groups_high};
\num[spts2]{fu.own.stack_over_low} to \num[spts2]{fu.own.stack_over_high}), above it on \num[n]{fu.own.stack_above}
benchmarks and below it on \num[n]{fu.own.stack_below}. Against the reference chosen by the own metric it is above
on \num[n]{fu.own.stack_above_own} benchmarks and below on \num[n]{fu.own.stack_below_own}, yet
\num[pts2]{fu.own.stack_minus_own} points below on average, because every fitted policy optimises log loss: on
iSarcasmEval stacking reaches an F\literal{1} of \num[pct1]{fu.own.isarcasm_stack}, the model chosen by the own
metric \num[pct1]{fu.own.isarcasm_own_reference}.

\subsection{Redrawn Folds}
\label{sec:further-checks}

\paragraph{Redrawn folds.}
The corpus has one assignment of source groups to folds, which our earlier analyses also used. We drew
\num[n]{folds.draws} new assignments, each dealing the shuffled source groups of a benchmark into five folds, and
refitted every method that the simulator fits (Table~\ref{tab:fold-redraws}). All
\num[n]{folds.suites.main35.contrasts_compared} contrasts of Table~\ref{tab:contrasts} between these methods keep
their sign on every draw, and no method moves by more than \num[f3]{folds.suites.main35.largest_log_loss_shift}
nats. The frozen folds do not favour the fitted combinations: stacking with a final temperature leads
$\mathcal{B}_d$ by \num[pts2]{fu.folds.C5_frozen} points on the frozen folds and by \num[pts2]{fu.folds.C5_mean} on
average over the draws, and greedy selection by \num[pts2]{fu.folds.D1_frozen} and \num[pts2]{fu.folds.D1_mean}.
The router trails it by \num[pts2]{fu.folds.C8_frozen} and \num[pts2]{fu.folds.C8_mean}.

\subsection{Further Analyses}
\label{sec:further-analyses}

\begin{table*}[tp]
\centering\scriptsize\setlength{\tabcolsep}{3pt}
\begin{tabular}{@{}lrrrrr@{}}
\toprule
Policy & Log loss & Acc.\ (\%) & Own (\%) & Calls & USD/1k \\
\midrule
\multicolumn{6}{@{}l}{\textit{Told the benchmark}} \\
Best model per benchmark & 0.384 & 86.3 & 78.7 & 1.00 & 0.150 \\
Best model per benchmark, by own metric & 0.427 & 83.8 & 81.0 & 1.00 & 0.110 \\
Ridge router, text embeddings & 0.380 & 86.2 & 80.0 & 1.00 & 0.096 \\
Ridge router, text embeddings, no cap & 0.379 & 86.4 & 80.6 & 1.00 & 0.108 \\
kNN router, whole request & 0.394 & 85.7 & -- & -- & 0.090 \\
Top-$k$ pool, validated $k$ & 0.363 & 87.2 & -- & 2.99 & 0.419 \\
Pool with equal weight per provider & 0.506 & 83.1 & -- & 17.00 & 1.749 \\
\midrule
\multicolumn{6}{@{}l}{\textit{Not told the benchmark}} \\
One model for all benchmarks & 0.497 & 81.4 & 76.1 & 1.00 & 0.073 \\
kNN router, text embeddings & 0.412 & 85.5 & 80.1 & 1.00 & 0.081 \\
kNN router, request features & 0.440 & 84.0 & 77.5 & 1.00 & 0.127 \\
Ridge router, text embeddings & 0.442 & 84.3 & 77.6 & 1.00 & 0.098 \\
\bottomrule
\end{tabular}
\caption{Policies of the further analyses on the headline suite (benchmarks weighted equally). Routers are fitted inside each benchmark unless they are not told the benchmark; then one recalibration, one setting and one budget serve all benchmarks and the neighbours come from every benchmark. The ridge router regresses each model's log score on the request's sentence embedding. Routers are shown under the cost cap unless marked; the routers that are not told the benchmark are shown without it. The top-$k$ pool chooses $k$ on the validation fold. --: not computed.}
\label{tab:followup-policies}
\end{table*}

These analyses were added after the review of a draft (Appendix~\ref{sec:reproducibility}); Table~\ref{tab:followup-policies}
gives their policies.

\paragraph{Routers that are not told the benchmark.}
The nearest-neighbour router on sentence embeddings that is not told the benchmark reaches
\num[pct1]{fu.unknown.across_accuracy}, \num[pts2]{fu.unknown.across_minus_global.difference} points above the one
global model of that setting (\num[spts2]{fu.unknown.across_minus_global.over_low} to
\num[spts2]{fu.unknown.across_minus_global.over_high} when benchmarks are resampled) and
\num[pts2]{fu.unknown.across_minus_best.difference} points below $\mathcal{B}_d$, with an interval that excludes
zero when source groups are resampled and includes it when benchmarks are. Under the cost cap of the paper's router
it is \num[pts2]{fu.unknown.across_capped_minus_text.difference} points less accurate than the text router that is
told the benchmark (\num[spts2]{fu.unknown.across_capped_minus_text.over_low} to
\num[spts2]{fu.unknown.across_capped_minus_text.over_high} when benchmarks are resampled), with a log loss
\num[a3]{fu.unknown.across_capped_minus_text_log_loss} nats higher. Of its neighbours,
\num[pct0]{fu.unknown.same_benchmark_share} come from the request's own benchmark, against
\num[pct0]{fu.unknown.chance_share} for neighbours drawn at random.

\paragraph{A ridge router.}
The ridge router inside each benchmark is as accurate as $\mathcal{B}_d$ under the cap
(\num[spts2]{fu.unknown.ridge_minus_best.difference} points; \num[spts2]{fu.unknown.ridge_minus_best.over_low} to
\num[spts2]{fu.unknown.ridge_minus_best.over_high} when benchmarks are resampled) at
\num[pct0]{fu.unknown.ridge_cost_share} of its cost, and \num[pts2]{fu.unknown.ridge_minus_text.difference} points
above the nearest-neighbour router with the same features. We ran \num[n]{fu.ridge.variants} variants of it, with or
without the cap and with either text representation, each once. All cost less than $\mathcal{B}_d$ (at most
\num[pct0]{fu.ridge.largest_cost_share} of its cost), and all have a lower log loss when source groups are
resampled; the \num[n]{fu.ridge.whole_request_variants} on the whole request also when benchmarks are. Under the cap
that variant is ahead by \num[a4]{fu.full_text.ridge_minus_best_log_loss.difference} nats
(\num[s4]{fu.full_text.ridge_minus_best_log_loss.within_low} to
\num[s4]{fu.full_text.ridge_minus_best_log_loss.within_high};
\num[s4]{fu.full_text.ridge_minus_best_log_loss.over_low} to
\num[s4]{fu.full_text.ridge_minus_best_log_loss.over_high}), and by
\num[a4]{fu.full_text.ridge_minus_best_log_loss_no_recall_only.difference} nats, with an interval that includes
zero, without the two recall-only benchmarks. The two variants without the cap are also more accurate than
$\mathcal{B}_d$ when source groups are resampled, though narrowly, with the stored text by
\num[pts2]{fu.unknown.ridge_uncapped_minus_best.difference} points
(\num[spts2]{fu.unknown.ridge_uncapped_minus_best.within_low} to
\num[spts2]{fu.unknown.ridge_uncapped_minus_best.within_high}); no variant is when benchmarks are resampled
(\num[spts2]{fu.unknown.ridge_uncapped_minus_best.over_low} to
\num[spts2]{fu.unknown.ridge_uncapped_minus_best.over_high} for the same one). Against the control the capped
router is \num[pts2]{fu.control.ridge_minus_control.difference} points ahead
(\num[spts2]{fu.control.ridge_minus_control.over_low} to \num[spts2]{fu.control.ridge_minus_control.over_high}),
with a lower log loss on \num[n]{fu.control.log_loss_better} benchmarks and a higher one on
\num[n]{fu.control.log_loss_worse}; in \num[n]{fu.control.cells_one_model} of the \num[n]{fu.control.cells}
combinations of benchmark and test fold it sends every request to one model.

\paragraph{The whole request.}
With the stored text representation, \num[n]{fu.full_text.stored_one_vector} benchmarks have one vector for every
request and \num[n]{fu.full_text.stored_below_half} have distinct vectors for fewer than half of their requests;
embedding the whole request leaves no benchmark below one half. The accuracy of the nearest-neighbour router does
not move (\num[spts2]{fu.full_text.knn_minus_stored.difference} points,
\num[spts2]{fu.full_text.knn_minus_stored.within_low} to \num[spts2]{fu.full_text.knn_minus_stored.within_high};
\num[spts2]{fu.full_text.knn_minus_stored.over_low} to \num[spts2]{fu.full_text.knn_minus_stored.over_high}), and it
stays below $\mathcal{B}_d$ by \num[pts2]{fu.full_text.knn_minus_best.difference} points. The ridge router's
accuracy changes by less than a tenth of a point.

\paragraph{Pools.}
The equal-weight pool of the best-ranked models with its size chosen on the validation fold calls
\num[f1]{fu.pool.top_k_calls} models on average and is level with stacking in accuracy
(\num[spts2]{fu.pool.top_k_minus_stacking.difference} points, \num[spts2]{fu.pool.top_k_minus_stacking.over_low} to
\num[spts2]{fu.pool.top_k_minus_stacking.over_high}), with a log loss \num[a3]{fu.pool.top_k_minus_stacking_log_loss}
nats higher. Weighting the providers equally instead of the models costs the equal-weight pool
\num[pts2]{fu.pool.balanced_minus_uniform.difference} points.
\section{Reproducibility}
\label{sec:reproducibility}

\paragraph{Code and data.}
The release has two forms. A Git repository (\url{https://github.com/Seffi-Robust-AI-Lab/DecisionReplay}) holds the code, the settings, the result files and the corpus in the layout that the simulator reads. A dataset on the Hugging Face Hub (\url{https://huggingface.co/datasets/seffico/DecisionReplay}) holds the same records as Parquet tables, with the benchmark and the model as columns, and they load with the \texttt{datasets} library.

\paragraph{Replay.}
Every result is computed by one command sequence from the released files and the labels rebuilt from the
benchmarks' own releases; no decision model was called, and two small local models computed the request
representations. \num[n]{det.reruns} further complete runs gave result files identical to the stored ones
(\num[n]{det.files_compared} files each). Tests of the simulator check that shuffling the labels of the test fold
changes neither the predictions nor the calls, and that replacing the outputs of uncalled models by missing values
changes nothing. All settings (grids, seeds, suites, contrasts) are in one file whose hash is written into every
result file. A full run takes \num[n]{timing.seconds} seconds with \num[n]{timing.processes} processes on a laptop
with \num[n]{timing.cpu_count} cores (Python \num[text]{timing.python}, NumPy \num[text]{timing.numpy}, SciPy
\num[text]{timing.scipy}, pandas \num[text]{timing.pandas}, PyArrow \num[text]{timing.pyarrow}, scikit-learn
\num[text]{timing.scikit_learn}).

\paragraph{Planned and exploratory analyses.}
The settings file fixed nine contrasts, their adjustment, the suite of benchmarks and the label budgets before the
first run. That run was on a larger collection, of which this corpus is the part that we release
(Appendix~\ref{sec:rights}), and two things changed afterwards. The models, the benchmark and the
\num[n]{rel.excluded_requests} requests that are left out were removed for their terms, after results had been seen:
the models and the benchmark after the results on the whole collection, and the requests after every analysis of
this paper had been run once with them. And two of the nine contrasts
concern a method whose weights earlier code had fitted on the whole collection; it cannot be replayed on the
released models, so \num[n]{facts.planned_contrasts.computed} contrasts remain (C\literal{1} to C\literal{5},
C\literal{8} and C\literal{9}), adjusted as a family of that size. The folds are not fresh either: our earlier
analyses used the same assignment (Appendix~\ref{sec:further-checks} redraws it). The paper therefore reports a
benchmark, not confirmatory tests.

Everything else was added after results had been seen and is descriptive: the router's per-benchmark control, the
contrasts D\literal{1} to D\literal{5}, the operating points without the cost cap, the further request
representations, the intervals that resample benchmarks, the metrics of LLMRouterBench, the hardness analysis and the
redrawn folds. The analyses of Appendix~\ref{sec:further-analyses} were added after a review of a draft; they come
from separate commands, each run once, and the whole-request representation from one run of the local sentence
encoder. The supplementary material lists these additions.
\end{document}